\documentclass[runningheads]{llncs}
\usepackage[T1]{fontenc}
\usepackage{graphicx}
\usepackage{breakcites}
\usepackage[colorlinks=true,breaklinks=true]{hyperref}
\usepackage{url}
\usepackage{color}

\usepackage{booktabs}   %% For formal tables:
\usepackage{subcaption} %% For complex figures with subfigures/subcaptions
\usepackage{listings}
\usepackage{relsize}
\usepackage{enumerate}
\usepackage{float}
\usepackage{nicefrac}
\usepackage{multirow}
\usepackage{microtype}
\usepackage{wrapfig}
\usepackage{amsmath}
\usepackage{amssymb}
\usepackage{mathtools}
\usepackage{xcolor}
\usepackage[algosection,ruled,vlined,linesnumbered,algo2e]{algorithm2e}
\usepackage{wrapfig}
\usepackage{algorithmic}
\usepackage[framemethod=tikz]{mdframed}
\usepackage{gensymb}
\usepackage[tableposition=below]{caption}
\makeatletter
\renewcommand\paragraph{\@startsection{paragraph}{4}{\z@}%
                       {3\p@ \@plus 0\p@ \@minus 0\p@}%
                       {-0.5em \@plus -0.22em \@minus -0.1em}%
                        {\normalfont\normalsize\bfseries}}
\makeatother

\definecolor{blue2}{rgb}{0.0, 0.5, 1.0}

\newcommand{\staterep}{\texttt{rep}}

\newcommand{\img}{\texttt{img}}
\newcommand{\Img}{\texttt{Img}}
\newcommand{\repImg}{\overline{\texttt{Img}}}
\newcommand{\cte}{\texttt{CTE}}
\newcommand{\he}{\texttt{HE}}
\newcommand{\vcte}{\texttt{cte}}
\newcommand{\vhe}{\texttt{he}}
\newcommand{\vctec}{\underline{\texttt{cte}}}
\newcommand{\vhec}{\underline{\texttt{he}}}
\newcommand{\checkdnn}{\texttt{check}}

\newcommand{\storm}{\textsc{Storm}}
\newcommand{\prism}{\textsc{PRISM}}
\newcommand{\fact}{\textsc{FACT}}
\newcommand{\deepdecs}{\textsc{DeepDECS}}

\definecolor{prismgreen}{rgb}{0, 0.6, 0}

\lstdefinelanguage{Prism}{ % syntax highlight via font
basicstyle=\color{red}\scriptsize\ttfamily, % small true type font (like courier)
keywords={bool,C,ceil,const,ctmc,double,dtmc,endinit,endmodule,endrewards,endsystem,F,false,floor,formula,G,global,I,init,int,label,max,mdp,min,module,nondeterministic,P,Pmin,Pmax,prob,probabilistic,R,rate,rewards,Rmin,Rmax,S,stochastic,system,true,U,X},
keywordstyle={\bfseries\color{black}},
numberstyle=\tiny\color{black},
comment=[l] {//}, morecomment=[s]{/*}{*/}, % single and multi-line
commentstyle= \color{prismgreen}, % dark green
tabsize=4, % tab treatment (going to be fixed in Prism)
captionpos=b, % put captions at the bottom
escapechar=@, % write LaTeX comments escaped by @ symbol
literate={->}{$\rightarrow{}$}{1}
}

\begin{document}
\title{
%Compositional Safety Proofs of Vision-based Discrete Controllers via Statistical Abstractions {\em or} 
%Probabilistic Analysis of Autonomous Systems with Machine Learning Perception}
%Closed-Loop 
%Safety Analysis of Autonomous Systems with Learning-enabled Visual Perception
%Compositional 
Closed-loop Analysis of Vision-based \\Autonomous Systems: A Case Study}
%-- remove for now for internal submission
%}
%
%\titlerunning{Probabilistic Safety Analysis of Autonomous Systems}
% If the paper title is too long for the running head, you can set
% an abbreviated paper title here
%
%updated for internal reiew
%to reflect contributions to discuss
\author{
Corina~P\u{a}s\u{a}reanu\inst{1,2}
\and
Ravi~Mangal\inst{2}
\and
Divya~Gopinath\inst{1}
\and
Sinem~Getir~Yaman\inst{3}
\and
Calum~Imrie\inst{3}
\and
Radu C\u{a}linescu\inst{3}
\and
Huafeng Yu\inst{4}
}

\authorrunning{C P\u{a}s\u{a}reanu et al.}
% First names are abbreviated in the running head.
% If there are more than two authors, 'et al.' is used.

\institute{
KBR, NASA Ames, Moffett Field CA 94035, USA\\
\and Carnegie Mellon University, Moffett Field CA 94035, USA\\
\and University of York, York, UK\\
\and Boeing Research and Technology, Santa Clara, CA\\
%\email{\{radu.calinescu,calum.imrie,sinem.getir.yaman\}@york.ac.uk}\\
%\email{calum.imrie@york.ac.uk}
%\email{divya.gopinath@nasa.gov}\\
%\email{\{rmangal,pcorina\}@andrew.cmu.edu}}
%\email{pcorina@andrew.cmu.edu}
%\email{sinem.getir.yaman@york.ac.uk}}
%
}
\maketitle              % typeset the header of the contribution
\begin{abstract}
Deep neural networks (DNNs) are increasingly used in safety-critical autonomous systems as  perception components processing high-dimensional image data.
Formal analysis of these systems is particularly challenging due to the complexity 
of the perception DNNs, the sensors (cameras), and the environment conditions. We present a case study applying formal probabilistic analysis techniques to an experimental autonomous system that guides airplanes on taxiways using a perception DNN. 
We address the above challenges by replacing the camera and the network with a compact probabilistic abstraction built from the confusion matrices computed for the DNN on a representative image data set. 
We also show how to leverage local, DNN-specific analyses as run-time guards to increase the safety of the overall system. 
%The approach 
%is compositional, as the abstraction is built separately from the rest of the system, and  
%enables scalable analysis with probabilistic guarantees for system-level safety properties.  
%Our approach enables the analysis of safety properties for autonomous systems that use complex DNNs for perception. %I cut scalable
%We verify the abstracted system using probabilistic model checking and confidence interval analysis and obtain probabilistic guarantees for system-level safety properties. 
Our findings are applicable to other autonomous systems that use complex DNNs for perception.

\end{abstract}
\setcounter{footnote}{0} 

\section{Introduction}
\label{sec:intro}

%%corina:
%-State the assumptions: we only consider classifiers; why is it realistic? cite ACASX which uses only output classes
%-why? simpler setting (robustness and calibration)
%-discrete-time models
%-deterministic environment

Complex autonomous systems, such as autonomous aircraft taxiing systems~\cite{KadronGPY21} and autonomous cars~\cite{abs-1910-07738,abs-1904-00649,huang2020survey}, need to perceive and reason about their environments using high-dimensional data streams (such as images) generated by rich sensors (such as cameras). Machine learnt components, specially deep neural networks (DNNs), are particularly capable of the required high-dimensional reasoning and hence, are increasingly used for perception in these systems. While formal analysis of the safety of these systems is highly desirable due to their safety-critical operational settings and the error-prone nature of learned components, in practice this is very challenging because of the complexity of the system components, including the high complexity of the neural networks (which may have thousands or millions of parameters), the complexity of the high-definition cameras that are used to capture the images, and the complexity of the environment in which the system operates (i.e., the world itself).

In this work, we describe a formal analysis of a closed-loop autonomous system that addresses the above challenges. 
%We demonstrate our analysis technique by means of a case study, that we use as a running example throughout the paper. 
Our case study is motivated by a real-world application, namely, an experimental autonomous system for guiding airplanes on taxiways developed by Boeing~\cite{ACT1,ACT2}. 

The key idea is to abstract away altogether the perception components, namely, the perception network and the image generator, i.e., the camera taking images of the world, and replace them with a probabilistic component $\alpha$ that maps (abstractions of) the state of the world to state estimates that are used in downstream decision making in the closed-loop system. The resulting system can then be analyzed with standard (probabilistic) model checkers, such as \prism{}~ \cite{PRISM} or \storm{}~\cite{STORM}.

The approach is {\em compositional}, in the sense that the probabilistic component is computed separately from the rest of the system. The transition probabilities in $\alpha$ are derived based on
%from the {\em accuracy} and 
{\em confusion matrices} computed for  the DNN (measured on representative data sets).
%; both are standard metrics for measuring DNN performance, promising easy adoption of our approach in practice. 
Developers routinely use confusion matrices to evaluate machine learning models, so our analysis is closely aligned with existing work-flows, facilitating its adoption in practice.

The size of the probabilistic abstraction  $\alpha$ is linear in the size of the output of the DNN, and is independent of the number of the DNN parameters or the complexity of the camera and the environment. 
We also describe how to leverage additional results obtained from analyzing the DNN in isolation (using DNN-specific methods \cite{ASE19,HuangKWW16,abs-1710-00486,GuoPSW17,KatzHIJLLSTWZDK19,leino21gloro}) to further refine the abstraction and also increase the safety of the closed-loop system through {\em run-time guards}.

The probabilities in $\alpha$ are estimated based on empirical data, so they are subject to error. We explore the use of {\em confidence intervals} in addition to point estimates for these probabilities and thereby strengthen the soundness of the analysis~\cite{calinescu2015formal,calinescu2016fact}. 
We also leverage rules %abstracting input-output behavior of 
mined from the DNN model \cite{ASE19} to act as run-time guards for the closed-loop analysis, thereby demonstrating its application in a realistic setting. %In order to also demonstrate the effect of some popular DNN-specific analysis methods, such as robustness certification \cite{} and confidence calibration, to serve as run-time guards, we also perform a second case study, where we build a simpler classifier model for the line-following robot problem. 
Our technique is applicable to other autonomous systems that use DNN-based perception from high-dimensional data. 

\paragraph{Related Work.}
Formal proofs of closed-loop safety have been obtained for systems with low-dimensional sensor readings ~\cite{santa2022nnlander,ivanov2021compositional,ivanov2019verisig,ivanov2020verifying,ivanov2021verisig,dawson2022learning,dawson2022safe}; however, they become intractable for systems that use rich sensors producing high-dimensional inputs such as images. 

%This is due to the difficulty of precisely modeling the camera~\cite{santa2022nnlander}, the external environmental conditions that are difficult to account for. Second, even if one manages to construct a mathematical model of the image generator (i.e., camera),  the combined complexity of it and the neural networks used for perception makes current closed-loop analysis algorithms prohibitively expensive. For instance, 

%NNLander-VeriF~\cite{santa2022nnlander} builds mathematical models of the camera and combines them with the neural network and the rest of the system, but can only handle small images and networks.
%black and white images with $16\times 16$ pixels and small networks.  
Other works  address the modeling and scalability challenges by constructing  {\em abstractions} of the perception components~\cite{hsieh2022verifying,katz2022verification}.  %To ensure soundness of the analysis, these abstractions need to be over-approximate but this leads to an accumulation of approximation error as safety is verified over longer horizons. Consequently, existing techniques either forego any soundness guarantees about their abstractions~\cite{katz2022verification,santa2022nnlander} or  provide a probabilistic guarantees of soundness for the used abstraction~\cite{hsieh2022verifying}; however these are not probabilistic guarantees over teh safety of the system.
To model different environment conditions, these abstract models use {\em non-deterministic} transitions.
%,  encoding the fact that a single underlying system state can lead to multiple different sensor readings depending on the environmental conditions, and therefore, a {\em set} of possible perception outcomes.
The resulting closed-loop systems are analyzed with traditional (non-probabilistic) techniques.
The abstractions either lack soundness proofs~\cite{katz2022verification} or come with only probabilistic soundness guarantees~\cite{hsieh2022verifying} which do not translate into probabilistic guarantees over the safety of the overall system.
%An alternate approach for scaling closed-loop safety proofs is to construct inductive invariants in the form of barrier functions~\cite{alur2011formal,dawson2022learning,dawson2022safe,tong2022enforcing}, %One only needs to prove that, over a single step of the system dynamics, the invariant is inductive and it implies safety. 
%which again is challenging in high-dimensional settings. 
VerifAI~\cite{ghosh2021counterexample} can find counter-examples to system safety, but can not provide guarantees.

In contrast to previous work, we describe a formal analysis that is {\em probabilistic}, which we believe is natural since the camera images capturing the state of the world are subject to randomness due to the environment; further DNNs are learnt from data and are not guaranteed to be 100\% accurate.
%In contrast to previous work, %instead of attempting to approximate or model the mapping between a given state and controller inputs, 
%we describe an analysis that is {\em probabilistic}, which we believe 
%and we use a small probabilistic abstraction for the vision components. 
%network together with the camera. 
%We believe that this probabilistic view 
%is natural 
%by representing the uncertainity in the outputs of the vision-based perception module as a probabilistic automaton. We believe that this makes sense 
%since \corina{to discuss:} DNNs are learnt from data and are not guaranteed to be 100\% accurate; further the environment %from which the input images are captured with the camera 
%is subject to unexpected perturbations. 
%We model the abstracted autonomous systems as a Discrete Time Markov Chain (DTMC) \cite{} and use methods for analysis of DTMCs \cite{} to reason about system-level properties. We obtain probabilistic guarantees about the validity of the probabilistic checks, even though the abstraction probabilities are empirical estimates.
%Through a confidence interval analysis, we provide probabilistic guarantees about system-level safety.
%
%\todo{to refine}
%\ravi{probabilistic abstraction of both, camera and perception modules; update the reason why probabilistic abstraction makes sense - because the camera is itself a probabilistic component. It takes in a state produces a distribution over images: corina: addressed but maybe not?}
%\corina{add more related work}
%For this case-study paper, 
We build on our previous work \deepdecs{}~\cite{calinescu2022discrete}, where 
%we %explored the 
%used confusion matrices to build abstract probabilistic components modeling the perception in autonomous systems. 
the goal is to perform controller synthesis with safety guarantees, so the formalism is more involved. Furthermore, \deepdecs{} 
%does not discuss a run-time guard mechanism and 
does not consider confidence interval analysis, which 
%to provide probabilistic guarantees for the results; 
we explore here based on some of our other previous works~\cite{calinescu2015formal,calinescu2016fact}. 
%\divya{DeepDecs also does not consider the impact of run-time guards on closed-loop safety analysis.} 
%In a different prior work, 
We analyzed center-line tracking using TaxiNet in~\cite{KadronGPY21}. That work focuses on the analysis of the network and not on the overall  system.
%\divya{Maybe make a contributions/novelty list?}

%\ravi{Probably we should also discuss \cite{tong2022enforcing,dean2020robust} since they both are about constructing safe controllers for systems with vision-based perception}

%We present a simple DTMC model of the abstracted autonomous system. We evaluate the scalability of our analysis and the effectiveness of refining probabilistic abstractions. 

\section{Autonomous Center-line Tracking with TaxiNet}
\label{sec:taxinet}
%-the taxinet problem\\

%\paragraph{Autonomous Center-line Tracking.} 
%Center-line Tracking on taxiways is one of the the most important ground operations in an airport.  An airplane is required to follow the center-lines of these taxiways during taxiing. 
%For this task, 
Boeing is developing an experimental autonomous system for center-line tracking on taxiways in an airport. The system  uses a neural network called TaxiNet for perception. TaxiNet is designed to take a picture of the taxiway as input and return the plane’s position with respect to the center-line on the taxiway. It returns two outputs; cross track error ($\vcte$), which is the distance in meters of the plane from the center-line and heading error ($\vhe$), which is the angle in degrees of the plane with respect to the center-line. These outputs are fed to a controller which in turn manoeuvres the plane such that it remains close to the center of the taxiway. This forms a closed-loop system where the perception network continuously receives images as the plane moves on the taxiway. We use this system as a case study and also as a running example throughout the paper.
%TaxiNet has been trained and tested using the X-Plane simulator \cite{xplane}. 
%Though center-lines on the pavement of airport taxiways and taxiways have standardized shapes and colors, their visibility maybe poor for a number of reasons, including skid marks on the taxiways, poor lighting conditions, and bad weather.
%In this study, our goal is to analyze the {\em safety} of the overall closed-loop autonomous system, accounting for the different visibility conditions.

\begin{figure}[t]
\centering
\begin{minipage}{.5\textwidth}
  \centering
  \includegraphics[width=0.9\linewidth]{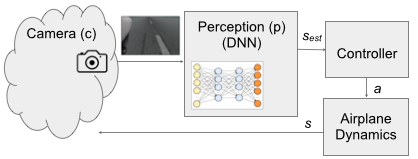}
  \captionof{figure}{Closed-loop System}
  \label{fig:auto_system}
\end{minipage}%
\begin{minipage}{.5\textwidth}
  \centering
  \includegraphics[width=0.9\linewidth]{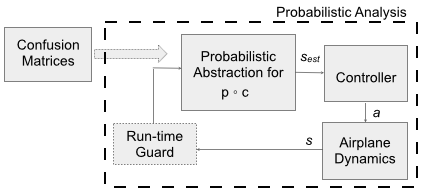}
  \captionof{figure}{Abstracted System}
  \label{fig:abs_system}
\end{minipage}
\vspace{-0.7cm}
\end{figure}

\paragraph{System Decomposition.} The decomposition of this system is illustrated in Figure~\ref{fig:auto_system}.
%
%The system consists of a {\em Controller}, the {\em Airplane} (which for simplicity we assume it also includes the Environment) and the {\em perception module} implemented as a DNN, i.e. TaxNet; there may be other sensors (radar, LIDAR, GPS) that for simplicity we abstract away here.  
%
The controller sends actions $a$ to the airplane to guide it on the taxiway. The dynamics (which models the movement of the airplane on the airport surface) maps previous state $s$ and action $a$ to the next state $s'$.\footnote{Velocity may be provided as feedback to the controller; we ignore here for simplicity.} 
Information about the taxiway is provided by the perception network ($p$), i.e. TaxiNet. The perception network takes  high-dimensional images captured with a {\em camera} ($c$), and returns its estimation $s_{est}$ of the real state $s$.
%\footnote{There may be other sensors (radar, LIDAR, GPS) that we abstract away for simplicity.} 

For our application, state $s \in S$ captures the position of the airplane on the surface; $S$ is modeled as $\cte \times \he$. The network estimates the state $s:=(\vcte,\vhe)$ based on images taken with a camera placed on the airplane. 
If the network is `perfect', then $s = s_{est}$.\footnote{Assuming the relevant state of the world is recoverable from the input image.} However, this does not hold in practice. The network is trained on a finite set of images and is not guaranteed to be 100\% accurate whereas images observed in operation show a wide variety due to different environment  (e.g., light, weather) conditions and imperfections in the camera.

%\begin{figure}[!t]
%\centering
%\includegraphics[scale=0.4]{images/ExampleImagesTaxiNet.png}
%\caption{Example input images for TaxiNet}
%\label{fig:images}
%\vspace{-0.25cm}
%\end{figure}

%{\bf moved from section 3 to be merged with text below?}

\paragraph{Component Modeling.}
%We built a simple discrete model of the aircraft dynamics; and a discretized controller for the system, ; the code is in the Appendix.

We built a simple discrete model of the airplane dynamics and  a discrete-time controller for the system, similar to previous related work \cite{HoffmannTMT07,479140} which also considers discretized control. Since the controller is discretized, we abstract the regression outputs of TaxiNet to view the model as a classifier which predicts the plane's position in discrete states. 
%\corina{maybe note similarity to acas-x here?} \divya{inputs to acasx are not images , does that matter?}%We hand-craft a discrete controller for the system which produces actions to manoeuvre the aircraft based on the predicted states. 
Treatment of more complex systems with continuous semantics and regression models is left for future work.
The main challenge that we address in the paper is the modeling of the perception components (the camera and the network), which we describe in detail in the next section.
%As mentioned, we use a probabilistic abstraction for the perception and the camera components of the system. We describe this abstraction in detail in the next section.
%
We model the (abstracted) autonomous system as a Discrete Time Markov Chain (DTMC)~\cite{DTMC}; the code is shown in the appendix. 

\paragraph{Safety Properties.}
In our study, the goal is to provide {\em guarantees} for safe behaviour with respect to two system-level properties indicated by our industrial partner.
The properties specify conditions for safe operation  in terms of allowed $\vcte$ and $\vhe$ values for the airplane, by using taxiway dimensions. 
%by using taxiway dimensions and have the following form $|\vcte|<8$ meters and $|\vhe|<35$ degrees. \ravi{I dont understand the previous sentence. What do 8 meters and 35 degrees represent?}
%\divya{we can perhaps just mention cte\_threshold and he\_threshold instead of 8 and 35?.}
The first property states that the airplane shall never leave the taxiway (i.e., $|\vcte|\leq8$ meters). The second property states that the airplane shall never turn more than a prescribed degree (i.e., $|\vhe|\leq35$ degrees), as it would be difficult to maneuver the airplane from that position.
%
%\emph{The goal of our analysis is to show that, 
%on a straight stretch of the taxiway, 
%the aircraft neither leaves the taxiway nor %becomes perpendicular 
%turns more than the prescribed angle from the center-line when the system is run for a fixed, finite number of steps}.
%and starts from a fixed initial state}. 
We encode the two %system-level safety 
properties 
%aim to use model checking to perform the analysis of the resulting system and check two system-level properties, which we write 
in PCTL \cite{Ciesinski2004} as follows.
$$P=? [ F (\vcte=-1)]\;\;\;\;\;\; \mathit{(Property\; 1)}$$
$$P=? [ F (\vhe=-1)]\;\;\;\;\;\;\; \mathit{(Property\; 2)}$$
%$$P=? [ F (pc=5~\&~v=1) ]~/~P=? [ %\neg(F (pc=5~\&~v=0)) ]$$
%$$P=? [F (pc=5)]$$
Here ``-1'' denotes an error for either $\vcte$ or $\vhe$ (meaning that the airplane is either off the taxiway or turned more than the prescribed angle) and $P=?$ indicates that we want to calculate the probability that eventually ($F$) the system reaches an error state. 

\section{Probabilistic Analysis}

%\begin{figure}[!t]
%\centering
%\includegraphics[scale=0.4]{abstraction.jpg}
%\caption{Probabilistic abstraction of $p\circ g$\todo{Improve figure; expand caption}}
%
%\label{fig:abstraction}
%\vspace{-0.25cm}
%\end{figure}

In this section,  we describe the methodology for abstracting and analyzing an autonomous system leveraging probabilistic model checking. 
%
%\subsection{Overview}
%{\bf moved from introduction}
%We are interested in closed-loop safety proofs that ensure safe behavior of the system over a finite planning horizon given a fixed initial state. 
%
%In this section, we present a probabilistic model checking approach for checking closed-loop safety of autonomous systems with discrete-time semantics.
%and deterministic components, by bringing together two ideas from our prior work~\cite{calinescu2015formal,calinescu2016fact,calinescu2022discrete}. 
%The challenge is to model the behaviour of the perception component, i.e., the neural network operating on high dimensional inputs, together with the camera in such a way that it becomes amenable for verification.
%The rest of the system (controller and dynamics) can be modeled either deterministically or probabilistically, following previous work; although very complex, we do not consider it as a challenge for this work.
%
The main idea, which we initially explored in \cite{calinescu2022discrete},
is to replace the composition $p \circ c$ of the camera (denoted as $c$) and the perception DNN (denoted as $p$) 
%First, we use the observation, proposed in \cite{calinescu2022discrete}, that 
%the composition of the generator and perception components (i.e., $p \circ g$ in Figure~\ref{fig:auto_system}) can be abstracted by 
with a probabilistic map $\alpha:S\to \mathcal{D}(S)$ from system states to a discrete distribution over system states. 
Figure~\ref{fig:abs_system} depicts the abstracted autonomous system. 

We observe that $c$ can be viewed as a map between state $s\in S$ to a distribution over images, denoted as $\mathcal{D}(\Img)$ , where $\img\in \Img$ and $\Img$ is the set of images. For instance, in the TaxiNet system, state $s$ only captures the position of the airplane with respect to the center-line, but there are many different images that correspond to the same position. This is due to uncontrollable environmental conditions, such as %markings on the runway/taxiway or 
temporary sensor failures or different lighting and weather conditions. 
Consequently, a single state $s$ can map to a number of different images depending on the environment, and this is modeled by considering $c$ to be a probabilistic map of type $S\to\mathcal{D}(\Img)$.
Given a system state $s$, $\alpha(s)$ models the probability of $p \circ c$ leading to a particular estimated state $s_{est}$;
%in the presence of environmental inputs drawn from some fixed distribution.
$\alpha$ needs to be probabilistic because $c$ itself is probabilistic and $p$ is not perfectly accurate.
The probabilities can be empirically estimated via a confusion matrix that records the performance of the network on a {\em representative} data set.

We further describe how we can leverage DNN-specific analysis to improve the accuracy of perception and the safety of the overall system, via the optional addition of run-time guards. 
For the verification of the closed-loop system, we use the \prism{} model checking tool \cite{PRISM}. We also explore methods for analysis of DTMCs with uncertain transition probabilities \cite{calinescu2015formal,calinescu2016fact}, to obtain {\em probabilistic guarantees} about the validity of our probabilistic safety proofs even though the abstraction probabilities are empirical estimates.

\paragraph{Assumptions.}
Our analysis assumes that the distribution of inputs to the network remains fixed over time (i.e., it is not subject to distribution shifts). 
%and there are no distribution shifts from when the representative data set of perception inputs is collected to when it s deployed. 
Moreover, the data set of input images used to estimate the probabilities in $\alpha$ is assumed to be \emph{representative}, i.e., constituted of independently drawn samples from this fixed underlying distribution of inputs.
Relaxing these assumptions is a challenging but important task for future research.

%\corina{say somewhere that this abstraction also has the advantage that it allows us to model a wide variety of environment conditions; random perturbations etc? }

%\begin{figure}[!t]
%\centering
%\includegraphics[scale=0.4]{images/abs_system2.png}
%\caption{Our approach}
%\caption{Abstracted autonomous system: $\nu$ probabilistically picks a partition $i$, $\alpha$ is a set of probabilistic maps $\{\alpha_1,\alpha_2,\ldots,\alpha_N\}$ from states to states and $i$ is the index of the map to be used.\todo{improve figure}}
%\label{fig:abs_system}
%\vspace{-0.25cm}
%\end{figure}

%\paragraph{Outline} In the following, we describe in more detail the probabilistic abstraction for perception; we then describe how we leverage DNN analysis as run-time guards to improve the accuracy of perception and improve  safety of the overall system; finally we describe the confidence interval analysis that is meant to provide the probabilistic guarantees.

\subsection{Probabilistic Abstractions for Perception}
\label{sec:abstractions}
%- Show a small figure(automaton)\\
%-Describe how we abstract the data generation and perception components\\
%-Explain why such an abstraction is valid; under no distribution shift assumption + confidence intervals\\
%-Explain how we abstract/update the DTMC using this abstraction\\
%-Theorem: If the system is proven safe using statistical abstraction, it is actually safe with a high probability \\
%- What do when there is zero evidence for a particular case?

%The main challenge that we address in this work is the verification of closed loop systems that integrate deep learning components that operate on high-dimensional data. We address this challenge with probabilistic abstractions, described next.

%{\bf  what are environmental inputs? images? in that case they are generated inputs}
%We begin our analysis by constructing a probabilistic abstraction $\alpha:S\to\mathcal{D}(S)$ of the composition  of the camera ($c$) and the perception network($p$)  ($p\circ c$) of the autonomous system. 
We describe in detail the construction of the probabilistic abstraction $\alpha:S\to\mathcal{D}(S)$.
%of the composition  of the camera ($c$) and the perception network($p$)  ($p\circ c$) of the autonomous system. 
We do not need access to the camera and only require black-box access to the network for constructing our abstraction.\footnote{Our run-time guard does require white-box access.}
%Given a state $s$, the probabilistic abstraction $\alpha$ returns a discrete distribution $\alpha(s)$ over states $S$. 
We assume $S$ is a finite set such that $\#S=K$ where $\#S$ denotes the cardinality of set $S$. We use $\alpha(s,s_{est})$ to represent the probability associated with estimated state $s_{est}$. It is defined as,
% \begin{equation}
%   \label{eq:alpha_def}
%   \alpha(s,s_{est}) := \Pr_{e\sim D}[(p\circ c)(s,e)=s_{est}]
% \end{equation}

% where $D$ is a fixed distribution that we assume the inputs are drawn from during the operation of the autonomous system. We define distribution $\overline{D}$ to be a distribution over the high-dimensional sensor readings that the perception component will encounter during its operation. We can define $\overline{D}$ using $D$ as follows,
% \begin{equation}
% \label{eq:overD_def}
% \Pr_{\overline{D}}[O=o] := \sum_{s\in S}\Pr_{e\sim D}[c(s,e)=o]
% \end{equation}
% where $\Pr_{\overline{D}}[O=o]$ is probability of a particular observation $o$ as per $\overline{D}$.
% In practice, $D$ and $\overline{D}$ are unknown and we empirically estimate the required probabilities.
\begin{equation}
  \label{eq:alpha_def}
  \alpha(s,s_{est}) := \Pr_{\img\sim c(s)}[p(\img)=s_{est}]
\end{equation}

% \begin{equation}
%   \label{eq:alpha_def}
%   \alpha(s,s_{est}) := \Pr_{o\sim D}[p(o)=s_{est} \wedge c^{-1}(o)=s]
% \end{equation}
% \begin{equation}
%   \label{eq:D_def}
% \Pr_{D}[O=o] := \sum_{s\in S}w(s)\cdot\Pr_{c(s)}[O=o]
% \end{equation}
% where $\Pr_{D}[O=o]$ is probability of a particular observation $o$ as per $D$ and $w:S\to[0,1]$ such that $\sum_{s\in S}w(s)=1$.
%[Prove $Pr_{o\sim D}[\phi(o) \wedge gt(o)=s] = Pr_{o\sim c(s)}[\phi(o)]$]

We estimate the probabilities in $\alpha$ by means of a confusion matrix.
Let  $\repImg_s \subseteq \Img$ denote a \emph{representative test dataset} for images corresponding to state $s$, i.e., every sample in $\repImg_s$ is assumed to be an independently drawn sample from $c(s)$. We assume access to representative test datasets corresponding to every state $s \in S$. Let $\repImg := \bigcup_{s\in S} \repImg_s$.
For any test input $\img\in \repImg$, let $p^*(\img)\in S$ be the label (i.e., the true underlying state) of $\img$, which is known since $\repImg$ is a test dataset.  For the sake of technical presentation, we assume a bijective map $\staterep:S\to [K]$ that maps every state in $S$ to a number in $[K]:=\{1,2,\ldots,K\}$.
We evaluate $p$ on the test dataset $\repImg$ to construct a $K\times K$ confusion matrix $\mathcal{C}$ such that, for any $k,k'\in [K]$, the element in row $k$ and column $k'$ of this matrix is given by the number of inputs from $\repImg$ with true state $\staterep^{-1}(k)$ that the perception network $p$ classifies as state $\staterep^{-1}(k')$.
\begin{equation}
  \label{eq:confusion}
  \mathcal{C}
  %{v_1,v_2,\ldots,v_n}
  [k,k'] := \#\left\{\img\in \repImg
  %{v_1,v_2,\ldots,v_n} 
  \mid  p^*(\img)=\staterep^{-1}(k) \wedge p(\img)=\staterep^{-1}(k')\right\}
\end{equation}

Given the confusion matrix $\mathcal{C}$, empirical estimates for the probabilities in $\alpha$ are calculated as follows, 
\begin{equation}
    \label{eq:alpha_probabilities}
    \alpha(\staterep^{-1}(k),\staterep^{-1}(k')):= \frac{\mathcal{C}[k,k']}{\sum_{k''\in[K]} \mathcal{C}[k,k'']}.
\end{equation}
%\end{theorem}

%\begin{table}[ht!]
%    \centering
%    \caption{Confusion Matrix for CTE}
%    \begin{tabular}{|c|c|c|c|}
%    \hline
%    Total = 11108 & \textbf{0} & \textbf{1} & \textbf{2} \\
%    \hline
%    \textbf{0} & 6789 & 17 & 248 \\
%    \textbf{1} & 1180 & 552 & 0 \\
%    \textbf{2} & 924 & 0 & 1398 \\
%    \hline
%    \end{tabular}   
%    \label{tab:confusion}
%\end{table}

%\begin{table}[ht!]
%    \centering
\begin{wraptable}{r}{0.5\linewidth}%\vspace{-0.5in}
    \begin{tabular}{c|c|c|c|c|}
    \multicolumn{2}{c}{}&\multicolumn{3}{c}{Predicted}\\
    \cline{2-5}
    &Total = 11108 & \textbf{0} & \textbf{1} & \textbf{2} \\
    \cline{2-5}
    \multirow{3}{*}{Actual}&\textbf{0} & 4748 & 2139 & 148 \\
    &\textbf{1} & 91 & 2010 & 0 \\
    &\textbf{2} & 744 & 211 & 1017 \\
    \cline{2-5}
    \end{tabular}
    \caption{Confusion Matrix for $\vhe$}
    \label{tab:confusion}
    \vspace{-0.3in}
\end{wraptable} 
%\end{table}

%% from wikipedia
\paragraph{TaxiNet Example.}
%A confusion matrix, also known as an error matrix, is a specific table layout that allows visualization of the performance of an algorithm, typically a classifier. Each row of the matrix represents the instances in the actual class while each column represents the instances in the predicted class. Based on this matrix, one can compute {\em accuracy} and number of true/false positives/negatives, all standard measures for measuring the performance of a neural network.

%So $\alpha(CTE=0,CTE_{est}=0)=6789/(6789+17+248)=0.962$. Similarly, $\alpha(CTE=2,CTE_{est}=0)=924/(924+1398)=0.398$, etc.

For the Taxinet application, we construct two probabilistic maps, $\alpha_{\vcte}$ and $\alpha_{\vhe}$, corresponding to each of the state variables $\vcte$ and $\vhe$, using a representative test data set with 11108 samples.\footnote{To simplify the DTMCs, we model the updates to $\vcte$ and $\vhe$ as independent. For more precision,  we can compute confusion matrices and $\alpha$ for the pair $(\vcte,\vhe)$.} Thus, $\alpha_{\vcte}$ is of type $\cte \to \mathcal{D}(\cte)$ and $\alpha_{\vhe}$ is of type $\he \to \mathcal{D}(\he)$.
Table~\ref{tab:confusion} illustrates the confusion matrix for $\vhe$.
The mapping $\alpha_{\vhe}$ is computed in a straightforward way: $\alpha_{\vhe}(0,0)=4748/(4748+2139+148)=0.675$, giving the probability of estimating correctly that the value of $\vhe$ is zero.
Similarly, $\alpha_{\vhe}(1,0)=91/(91+2010)=0.043$, giving the probability of estimating incorrectly that the value  of $\vhe$ is zero instead of one.
%Accuracy is the number of correct predictions divided by total number of instances, i.e. $4748+2010+1017/11108=0.7$
The corresponding DTMC code is as follows:

%\begin{verbatim}
%[]cte=0 -> 0.962:(cte_est'=0) + 0.002:(cte_est'=1) + 0.036:(cte_est'=2);
%[]cte=1 -> 0.681:(cte_est'=0) + 0.319:(cte_est'=1);
%[]cte=2 -> 0.398:(cte_est'=0) + 0.602:(cte_est'=2);    
%\end{verbatim}

%\begin{verbatim}
\begin{lstlisting}[language={Prism}, rulesepcolor=\color{black}, rulecolor=\color{black}, breaklines=true, breakatwhitespace=true,frame=single]
[] he=0 -> 0.675: (he_est'=0) + 0.304: (he_est'=1) + 0.021: (he_est'=2);     
[] he=1 -> 0.043: (he_est'=0) + 0.957: (he_est'=1) + 0.0: (he_est'=2);
[] he=2 -> 0.377: (he_est'=0) + 0.107: (he_est'=1) + 0.516: (he_est'=2);
\end{lstlisting}
%\end{verbatim}       

%% add code for HE and perhaps a picture?

A similar computation is performed for constructing $\alpha_{\vcte}$. The resulting code for the closed-loop system is shown in Appendix~\ref{sec:dtmc_prism_no_vc}. 

%\subsection{Probabilistic Abstraction Refinement via Run-time Checks}
\subsection{DNN Checks as Run-Time Guards}
\label{sec:abs_refine}

%We discuss here how we can incorporate the results of local, DNN-specific checks into the overall analysis of the closed-loop system. %In the past few years, a plethora of techniques have been proposed that focus specifically on verification of DNNs. They verify properties such as {\em local robustness}
%\footnote{$p$ is locally robust at $\img$ if $\forall \img'\in \Img. ||\img'-\img||\leq\epsilon \implies p(\img)=p(\img')$}} 
%of DNNs to perturbations of individual inputs~\cite{} or more general safety properties of DNNs~\cite{Marabou}.
%However, it is unclear how the obtained proofs can be used to provide system-level guarantees for the end-to-end systems that use the verified DNN components.

%To address this serious limitation, 
We use
%experiment with the idea of leveraging 
DNN-specific checks as run-time guards to improve the performance of the perception network and therefore the safety of the overall system.  We hypothesize that for inputs where the checks pass, the network is more likely to be accurate, and therefore, the system is safer.

%A similar idea was first introduced 
 
%\deepdecs{} uses DNN checks to characterize the trustworthiness of the DNN prediction at an input and  synthesizes safer, more performant controllers by allowing the controller logic to depend on the results of these DNN checks. In contrast, we deploy DNN checks, independently of the controller, as run-time guards to improve the safety of the system. 
%\corina{to check if that paper talks about run-time guards; expand on difference; say we use patterns in addition to local robustness?}

%\footnote{Examples of pointwise DNN checks include local robustness verifiers~\cite{},
%out-of-distribution detectors~\cite{}, and activation pattern based checks~\cite{}.} 

%We assume that a system that uses pointwise DNN checks as run-time guards goes to a special $\abort$ (or fail-safe) state whenever any check fails; the logic can be generalized and consider other safe-mode operation. 
%For such systems, we can perform a \emph{conditional} safety analysis, i.e., we can analyze the safety of the system conditioned on it not entering the $\abort$ state. We hypothesize that for inputs where all the checks pass, the network is more likely to be accurate, and therefore, the system is safer.

For our case study, we extract input-output rules from the DNN 
%in terms of the internal neuron values of the network 
and use them as run-time guards (as described in Section~\ref{sec:eval}). 
%For the other case study, we consider popular point-wise DNN checks, such as local robustness; an image  classifier $p$ is locally robust at $\img$, if $\forall \img'\in \Img. ||\img'-\img||\leq\epsilon \implies p(\img)=p(\img')$, 
%for some distance metric such as $l_2$ or $l_\infty$ and a small $\epsilon$.
More generally, one can use any off-the-shelf pointwise DNN check, such as local robustness \cite{gowal2018effectiveness,raghunathan2018certified,singh2019abstract,pmlr-v97-cohen19c,fromherz2021fast,leino21gloro} or confidence checks for well-calibrated networks~\cite{GuoPSW17},  as run-time guards (provided that they are fast enough to be deployed in practice). For practical reasons (TaxiNet is a regression model, it contains ELU~\cite{clevert2015fast} activations, we do not have access to the training data) we can not use off-the-shelf  checks here. %\corina{please check above}\ravi{looks good}

%In our previous work \deepdecs{}~\cite{calinescu2022discrete}, we also explored the use  DNN-specific analysis to inform controller synthesis, but we present a different formulation here and design a run-time guard mechanism.

\paragraph{Modeling DNN Checks.} Let us denote the application of (one or more) DNN-specific checks as a function $\checkdnn:(\Img\to S) \times \Img\rightarrow \mathbb{B}$, such that, for perception network $p \in \Img\to S$ and image $\img\in \Img$, $\checkdnn(p,\img)=\mathsf{true}$ if $p$ passes the checks at input $\img$, and $\checkdnn(p,\img)=\mathsf{false}$ otherwise. 

%We model the result of applying one or more DNN checks with a binary variable $v$; $v=1$ if {\em all} checks return true for an image and $v=0$ otherwise.
%\footnote{We use a simpler, more scalable formulation than in~\cite{calinescu2022discrete}}
We further assume that a system that uses DNN checks as a run-time guard attempts to read the camera sensor multiple (one or more) times, until the check passes; %at the same time it counts the number of failed checks 
and aborts (or goes to a fail-safe state) if the number of consecutive failed checks reaches a certain threshold. %\corina{check}\ravi{looks good} 
This logic can be generalized to consider more sophisticated safe-mode operations; for instance, the system can decelerate and/or notify an operator when the threshold is reached, as this could indicate serious sensor failure or adverse weather conditions.
%slow down when the check fails or it can attempt to read the sensor multiple times, until the check passes or it can count the number of failed checks and notify an operator when a certain threshold is reached, etc. 
%For such systems, we can perform a \emph{conditional} safety analysis, i.e., we can analyze the safety of the system conditioned on it not entering the $\abort$ state. We hypothesize that for inputs where all the checks pass, the network is more likely to be accurate, and therefore, the system is safer.

To model the effect of the run-time check in our analysis, we can define $\beta$ as the probability that an image $\img$ generated by the camera $c$, for {\em any} state $s$, satisfies $\checkdnn(p,\img)=\mathsf{true}$;
\begin{equation}
  \label{eq:beta_def}
  \beta := \Pr_{\img\sim D}[\checkdnn(p,\img)=\mathsf{true}]
\end{equation}
\noindent Here $D$ is the distribution obtained by {\em combining}  $c(s)$ for all states $s\in S$.\footnote{To simplify the presentation, we omit the precise mathematical formulation for $D$.} To be more precise we can define a separate $\beta_s$ for each state $s$. We estimate $\beta$ using the representative set of images
%
%can partition the input space {\tt Img}  of the perception component into two sets based on the value of $v$ (0 or 1). We define $\beta$ as the probability that, for a uniformly sampled state $s$, the image generated by the camera $c$ belongs to the partition with $v=1$. $\beta$ can be calculated as follows,
%\begin{equation}
%  \label{eq:beta_def}
%  \beta := \frac{1}{\# S}\sum_{s\in S}\Pr_{\img\sim c(s)}[\checkdnn_1(p,\img)=\mathsf{true}~\wedge~\ldots~\wedge \checkdnn_N(p,\img)=\mathsf{true} ]
%\end{equation}
%where $\checkdnn_1,\ldots,\checkdnn_N$ are the $N$ DNN checks that all need to pass in order for $v=1$ and $\# S$ is the cardinality of the finite set $S$ of states. This definition of $\beta$ assumes that each state $s \in S$ appears with equal frequency during the operation of the closed-loop system. To be more precise, one could define a separate parameter $\beta_s$ for every state $s \in S$.
%
%Given a representative test dataset 
$\repImg$, 
%we can calculate an estimate of $\beta$ as,
\begin{equation}
  \label{eq:beta_est}
  \beta := \frac{\# \repImg^{true}}{\# \repImg}
\end{equation}
where $\repImg^{true}:=\{\img\in \repImg \mid  \checkdnn(p,\img)=\mathsf{true}\}$.
%$\repImg^v:=\left\{\img\in \repImg
%  \mid  \checkdnn_1(p,\img)=\mathsf{true}~\wedge~\ldots~\wedge~\checkdnn_N(p,\img)=\mathsf{true}\right\}$.

For the overall analysis of the closed-loop system, irrespective of the state $s$, we can assume that the DNN check will pass with a probability $\beta$. Moreover, 
%since the system transitions to an $\abort$ state for all the images where the check fails, 
since the perception network only processes images that pass the DNN check, we construct a refined probabilistic abstraction $\alpha^{true}$ using conditional probability: 
\begin{equation}
  \label{eq:alphav_def}
  \alpha^{true}(s,s_{est}) := \Pr_{\img\sim c(s)}[p(\img)=s_{est}|\checkdnn(p,\img)=\mathsf{true}]
  %~\wedge~\ldots~\wedge~\checkdnn_N(p,\img)
  %=\mathsf{true}]
\end{equation}

We can estimate $\alpha^{true}$ as before, but the confusion matrix is built using only the images that pass the DNN check, i.e., for dataset $\repImg^{true}\subseteq\repImg$. 
%In other words, $\alpha^v$ is estimated as described in Section~\ref{sec:abstractions} using the confusion matrix constructed for dataset $\repImg^v\subseteq\repImg$.

%The refined abstraction is a pair $(\alpha,\nu)$. $\alpha$ is a set of probabilistic maps, $\{\alpha_1, \alpha_2,\ldots,\alpha_N\}$, one for each partition, and $\nu$ is a map from states $S$ to a discrete distribution over the partition indices $\{1,2,\ldots,N\}$. Given a state $s'$, it models the probability that the sensor reading generated by $g$ for $s'$ and random environmental input $e$ is a member of a particular partition. 
%For instance, local robustness\footnote{$p$ is locally robust at $\img$ if $\forall \img'\in \Img. ||\img'-\img||\leq\epsilon \implies p(\img)=p(\img')$} is an input-specific, run-time checkable property that yields two partitions of the input space.

%{\bf mention here the property about how often the system abstains as well as the property with conditional probabilities?}

%\paragraph{Candidate DNN Checks as Run-time Guards.} 
\paragraph{TaxiNet Example.} 
%An image  classifier $p$ is \emph{locally robust} at $\img$, if $\forall \img'\in \Img. ||\img'-\img||\leq\epsilon \implies p(\img)=p(\img')$, 
%for some distance metric such as $l_2$ or $l_\infty$ and a small $\epsilon$. A classifier $p$, that first assigns a probability to every label and then outputs the label with the highest probability $\gamma$ as its prediction, is \emph{confident} at an input $\img$ if $p$ is well-calibrated~\cite{} and $\gamma$ (i.e., probability of the predicted class) at $\img$ is greater than a predefined threshold.
%\corina{to expand: by adding calibration}
%\paragraph{TaxiNet Example.} 
%Going back to our example example, 
For TaxiNet, out of 11108 inputs, 9125 inputs (i.e., 82.1\%) pass the DNN check
%(rules in terms of neuron values characterizing misbehavior) 
resulting in the following code:

%\begin{verbatim}
\begin{lstlisting}[language={Prism}, rulesepcolor=\color{black}, rulecolor=\color{black}, breaklines=true, breakatwhitespace=true,frame=single]
i:[0..M] init 0; 
[]  pc=0 & i<M -> 0.821: (v'=1) & (pc'=1) & (i'=0) + 0.179: (v'=0) & (i'=i+1);
\end{lstlisting}
%\end{verbatim}

\noindent We model the result of applying the DNN check with variable $v$; $v=1$ if the check returns $\mathsf{true}$ for an image and $v=0$ otherwise. $M$ is the number of allowed repeated sensor readings and $i$ is used to count the number of failed DNN checks.

The abstraction for state variables $\vhe$ ($\alpha_{\vhe}$) and $\vcte$ ($\alpha_{\vcte}$) is only computed for the inputs that pass the check (i.e., for $v=1$) based on newly computed confusion matrices.
The DTMC code for the closed-loop system with run-time guards is in Appendix~\ref{sec:dtmc_prism_vc}.

\subsection{Confidence Analysis}
\label{sec:conf_analysis}
%\textcolor{red}{refer to experiments: describe the stages of \fact{}, explain the CI analysis from Table2}
The construction of the probabilistic abstractions %described in the previous sections 
relies on calculating empirical point estimates of the required probabilities. 
However, these empirical estimates lack statistical guarantees and can be off by an arbitrary amount from the true probabilities. 
%The probabilistic verification can propagate and amplify these errors, leading to analysis results that can not be trusted. 
To address this concern, we experiment with using \fact{}~\cite{calinescu2015formal,calinescu2016fact} to calculate {\em confidence intervals} 
%for each estimate at a user-defined confidence threshold and use these for establishing confidence intervals 
for the probability that the safety properties of the closed-loop system are satisfied.
%
%Concretely, we incorporate these confidence intervals into the DTMC model of our abstract autonomous system. The probabilities of the transitions corresponding to the probabilistic abstractions are no longer described by single numbers but instead by confidence intervals. To analyze such parametric DTMCs, we use the \fact{} method described in \cite{calinescu2015formal,calinescu2016fact}. 
%
%- Give a brief technical description of the \fact{} approach by Calinescu et al. and how we apply it to our setting
%
%copied from Radu's paper
%\fact{} (Formal verificAtion with Confidence inTervals) analyzes DTMCs with unknown state transition probabilities, when observations of these transitions are available from logs or run-time observations of the modelled system. 
%Given a property of the modelled software system that is formally expressed in probabilistic temporal logic, and a confidence level $1-\alpha\in(0,1)$ , \fact{} syntheses a $1-\alpha$ confidence interval for the property. 
%As a result, \fact{} enables the rigorous analysis of the safety (and other properties) of the systems when the transition probabilities between the states of the MCs used to model these systems are approximated by sets of observations.
%
The inputs to \fact{} are: 1) a parametric DTMC $m$ where each empirically estimated transition probability is represented by a parameter, 2) a PCTL formula $\phi$, 3) an error level $\delta\in(0,1)$ and 4) an {\em observation function} $O$ mapping state $s$ to a tuple representing the number of observations for each outgoing transition from $s$; in our case, the number of observations 
%for transitions in the probabilistic abstraction $\alpha$ 
can be obtained directly from the computed confusion matrices, i.e., $O(s)=(\mathcal{C}[\staterep(s),1], \ldots, \mathcal{C}[\staterep(s),K] )$.
%
%The inputs to \fact{} are as follows: 1) a parametric Markov Chain in extended version of \prism{} language with the number of observations $M$, 2) a PCTL formula $\phi$, 3) a range of confidence (e.g. $0.95-0.99$). %$\alpha\in(0,1)$ and 4) an {\em observation function} mapping a state $s_k$ to 
%
%For the analysis of property $\phi$, 
\fact{} synthesizes a 
$(1-\delta)$-confidence interval 
$[a, b]\subseteq[0,1]$ 
for the probability that $\phi$ is satisfied, given the observations.  

%The \fact{} synthesis 
%of $(1-\alpha)$ confidence intervals for $\phi$ 
%is performed in two stages: 1) {\em parametric model checking} is used to generate an algebraic expression for $\phi$, which is a multivariate rational function that depends on the unknown transition probabilities; 2) {\em confidence interval inference}  uses the observations $O$ to derive confidence intervals for each unknown transition probability appearing in the algebraic expression and then uses them to derive the required confidence interval for $\phi$; the analysis is based on the concept of simultaneous confidence intervals for a multinomial distribution and requires solving optimization problems. More details can be found in \cite{calinescu2015formal,calinescu2016fact}.

%\corina{- discuss scalability issues? mention the newer work that scales better?}
\paragraph{Taxinet Example.}
%\fact{} analyzer requires the number of observations for each outgoing probabilistic transition which can be obtained directly from the computed confusion matrices, i.e., $O(s_k)=\{\mathcal{C}[k,k'] | k'=1..K\}$. 
The following partial code illustrates the parametric version of the code provided in Section \ref{sec:abstractions} (with the complete code for the parametric models provided in Appendix~\ref{sec:dtmc_fact}). The first three lines represent the number of observations obtained from the confusion matrix in Table \ref{tab:confusion}. 
%, e.g., \corina{to fix} $O(he_0)=\{\mathcal{C}[0,0], C[0,1], C[0,2]\} = \{4748, 2139, 148\}$. 

%\corina{give concrete example for our analysis; illustrate the computation of one interval based on the confusion matrix?}

%\begin{verbatim}
\begin{lstlisting}[language={Prism}, rulesepcolor=\color{black}, rulecolor=\color{black}, breaklines=true, breakatwhitespace=true,frame=single]
param double x = 4748 2139 148;
param double y = 91 2010;
param double z = 744 211 1017;
...
[] he=0 -> x1:(he_est'=0) + x2:(he_est'=1) + (1-x1-x2):(he_est'=2);
[] he=1 -> y1:(he_est'=0) + (1-y1):(he_est'=1);
[] he=2 -> z1:(he_est'=0) + z2:(he_est'=1) + (1-z1-z2):(he_est'=2);
\end{lstlisting}

\section{Experiments}
\label{sec:eval}
%In this section we report on experiments demonstrating probabilistic safety analysis for the two case studies, center-line tracking and line-following robot.
In this section, we report on the experiments that we conducted as part of our probabilistic safety analysis of the center-line tracking autonomous system.

%\subsection{Setup}
%\label{sec:eval_setup}
We built two DTMC models, $m_1$ and $m_2$, denoting the closed-loop center-line tracking system without and with a run-time guard, respectively. The airplane dynamics and the controller are identically modeled in the two DTMCs as discrete components.  
The code for the models (in \prism{} syntax) and more details about the analysis are presented in the appendix.
%The DTMCs in \prism{} syntax are shown in Appendix~\ref{sec:dtmc_prism}. 

%This logic induces a partition on the DNN outputs that we describe below. The models could be refined by considering a more realistic, finer discretization for the logic of the controller and the dynamics (e.g., derived from simulations), which in turn would induce a finer partition for the DNN. However, in our case study we focus on the main challenge of reasoning about the complex DNN working side-by-side with the other, more traditional components, for which the current logic is sufficient. 
%The scenario considered was, starting from an initial position, the airplane moving forward in a straight line for a fixed number ($N$) of steps. 

%\subsection{Discretization}

%\paragraph{DNN  discretization.} 
\paragraph{TaxiNet DNN.} 
%The TaxiNet DNN used for perception 
This is a regression model with  24 layers including five convolution layers, and three dense layers (with 100/50/10 ELU neurons) before the output layer.
%\ravi{number of layers only adds up to 9; also we say 4 dense layer but give number of neurons for only 3 layers}
%\divya{The model has other layers like batch normalization apart from conv and dense layers. I counted the number of layers in the summary and it comes up to 21. However, in our earlier case study paper with Boeing, we have stated 24 layers. Maybe some layers shown in the summary are internally made up of more than one units}. 
The inputs to the model are RGB color images of size 360 × 200 pixels. We use a representative data set with 11108 images, shared by our industry partner. %, for evaluating TaxiNet. 
The model has a Mean Absolute Error (MAE) of 1.185 for $\vcte$ and 7.86 for $\vhe$ outputs respectively.  %(including images with different environment conditions such as lighting and weather conditions\divya{we do not have different weather conditions. but we do have images with shadow, skid marks, night-time, and cross-roads}),
%shadow, skid marks, intersections and night-time),
%provided by our industry partner. 
%The closed-loop system encoded as a DTMC is shown in the Appendix~\ref{sec:dtmc_prism}.
%
The discrete nature of the controller in our DTMCs induces a discretization on TaxiNet's outputs and the treatment of TaxiNet as a classifier for the purpose of our analysis.
%The controller induces a discretization on the network's outputs as follows.
%The $\vcte$ output is partitioned into into five bins (i.e., labels) as follows:
The $\vcte \in [-8.0~\text{m},8.0~\text{m}]$ and $\vhe \in [-35.0\degree,35.0\degree]$ outputs of the regression model are translated into classifier outputs $\vctec \in \{0,1,2,3,4\}$ and $\vhec \in \{0,1,2\}$ as shown below.

%\ravi{is it correct that the regression output for $\vcte$ is always between -8 m and 8 m?} \divya{In the Taxinet code, the model outputs cte and he are between 0 and 1. These are then scaled up to -8.0, 8.0 for cte and -35.0 , 35.0 for he}
%\begin{itemize}
% \item label "3" (\emph{far left} of center-line): -8.0 meters $<=$ $\vcte$ $<=$ -4.8 meters,
% \item label "1" (\emph{left} of center-line): -4.8 meters $<=$ $\vcte$ $<=$ -1.6 meters, 
% \item label "0" (\emph{on} center-line): -1.6 meters $<$ $\vcte$ $<=$ 1.6 meters, 
% \item label "2" (\emph{right} of center-line): 1.6 meters $<$ $\vcte$ $<=$ 4.8 meters,
% \item label "4" (\emph{far right} of center-line): 4.8 meters $<=$ $\vcte$ $<=$ 8.0 meters.
%\end{itemize}

% \begin{equation*}
%   \vctec =
%   \setlength{\arraycolsep}{0pt}
%   \renewcommand{\arraystretch}{1.2}
%   \left\{\begin{array}{l @{\quad} l @{\quad} l} 
%         3            & \text{if}~-8.0~\text{m}~ 
%         <= \vcte <= -4.8~\text{m} & \text{(\emph{far left} of center-line)} \\
%         1            & \text{if}~-4.8~\text{m}~ 
%         <= \vcte <= -1.6~\text{m} & \text{(\emph{left} of center-line)} \\
%         0            & \text{if}~-1.6~\text{m}~ 
%         <= \vcte <= 1.6~\text{m} & \text{(\emph{on} center-line)} \\
%         2            & \text{if}~1.6~\text{m}~ 
%         <= \vcte <= 4.8~\text{m} & \text{(\emph{right} of center-line)} \\
%         4            & \text{if}~4.8~\text{m}~ 
%         <= \vcte <= 8.0~\text{m} & \text{(\emph{far right} of center-line)}
%   \end{array}\right.
% \end{equation*}

\noindent\begin{minipage}{.5\linewidth}
\begin{equation*}
\resizebox{0.9\hsize}{!}{
  $\vctec =
  \setlength{\arraycolsep}{0pt}
  \renewcommand{\arraystretch}{1.2}
  \left\{\begin{array}{l @{~} l} 
        3            & \text{if}~-8.0~\text{m}~ 
        <= \vcte <= -4.8~\text{m} \\
        1            & \text{if}~-4.8~\text{m}~ 
        <= \vcte <= -1.6~\text{m}  \\
        0            & \text{if}~-1.6~\text{m}~ 
        <= \vcte <= 1.6~\text{m}  \\
        2            & \text{if}~1.6~\text{m}~ 
        <= \vcte <= 4.8~\text{m}  \\
        4            & \text{if}~4.8~\text{m}~ 
        <= \vcte <= 8.0~\text{m} 
  \end{array}\right.$}
\end{equation*}
\end{minipage}%
\begin{minipage}{.5\linewidth}
\begin{equation*}
\resizebox{0.9\hsize}{!}{
  $\vhec =
  \setlength{\arraycolsep}{0pt}
  \renewcommand{\arraystretch}{1.2}
  \left\{\begin{array}{l @{~} l} 
        1           & \text{if}~-35.0\degree~ 
        <= \vhe <= -11.67\degree \\
        0           & \text{if}~-11.67\degree~ 
        <= \vhe <= 11.66\degree \\
        2           & \text{if}~11.66\degree~ 
        <= \vhe <= 35.0\degree
  \end{array}\right.$}
\end{equation*}
\end{minipage}

% \noindent The $\vhe \in [-35.0~\text{degrees},35.0~\text{degrees}]$ output of the regression model is translated into classifier output $\vhec \in \{0,1,2\}$ as follows:
% \begin{itemize}
% \item label "1" (\emph{heading left}): -35.0 degrees $<=$ $\vhe$ $<=$ -11.67 degrees, 
% \item label "0" (\emph{heading straight}): -11.67 degrees $<$ $\vhe$ $<=$ 11.66 degrees, 
% \item label "2" (\emph{heading right}): 11.66 degrees $<$ $\vhe$ $<=$ 35.0 degrees.
% \end{itemize}

% \begin{equation*}
%   \vhec =
%   \setlength{\arraycolsep}{0pt}
%   \renewcommand{\arraystretch}{1.2}
%   \left\{\begin{array}{l @{\quad} l @{\quad} l} 
%         1           & \text{if}~-35.0~\text{degrees}~ 
%         <= \vhe <= -11.67~\text{degrees} & \text{(\emph{heading left})} \\
%         0           & \text{if}~-11.67~\text{degrees}~ 
%         <= \vhe <= 11.66~\text{degrees} & \text{(\emph{heading straight})} \\
%         2           & \text{if}~11.66~\text{degrees}~ 
%         <= \vhe <= 35.0~\text{degrees} & \text{(\emph{heading right})} \\

%   \end{array}\right.
% \end{equation*}

We use label ``-1'' to denote error states, i.e., $\vctec=-1$ iff $|\vcte|>8$ m and $\vhec=-1$ iff $|\vhe|>35\degree$. 
%\corina{maybe we should actually fix this notation} 
For simplicity, we use $\vcte$ and $\vhe$ to denote both the classifier and regression outputs in other parts of the paper (with meaning clear from context). Note that none of the input images are labeled by the classifier as ``-1'';
%, meaning that the network by itself can not lead to violations of the safety properties; 
however, this does not preclude the system from reaching an error.

\paragraph{Mining Rules for Run-time Guards.} 
We leverage our prior work~\cite{ASE19}, to extract rules of the form $Pre \implies Post$ from the DNN. %where $Pre$ is a condition on the internal neuron values of the DNN and $Post$ is a condition on the outputs of the DNN. 
%Such rules are interpreted to say that if the internal neuron values of the model generated in response to an input $\img$ satisfy $Pre$, then the output of the model on $\img$ is likely to satisfy $Post$. 
%
%For use as run-time guards, 
%We used a separate data set of 5554 input images to extract rules from the TaxiNet regression model such that 
$Post$ is the condition $|\vcte^{*}-\vcte| > 1.0~\text{m}\vee|\vhe^{*}-\vhe| > 5\degree$ on the regression model's outputs and $Pre$ is a condition over the neuron values in the three dense layers of TaxiNet ($\vcte^{*}$ and $\vhe^{*}$ denote ground-truth values). The considered $Post$ characterizes model misbehavior (as explained in \cite{KadronGPY21}).
%\ravi{I dont understand the comment that the same condition is used for computing MAE. Isn't MAE a measure for the regression model?}\corina{yes Post is computed on the regression model's output}
If an input satisfies $Pre$, the DNN check is considered to have failed on that input.  $Pre$ can be evaluated efficiently during the forward pass of the model, making it a good run-time guard candidate.
More details on the rules 
%and their deployment as runtime guards 
are in Appendix~\ref{sec:rules}.

\paragraph{Confusion Matrices.} 
%and Transition Probabilities.}
The confusion matrices for the classification version of TaxiNet, computed for the two cases (without and with run-time guard) 
%on the  inputs from the representative data set and on the inputs that pass the run-time check, respectively,   %(along with the estimated transition probabilities in the probabilistic abstraction $\alpha$) 
are shown in 
Tables~\ref{tab:trans_prob_5} and \ref{tab:trans_prob_vc_5} in Appendix~\ref{sec:tables}.
%\corina{from Huafeng} 
The tables can be used by developers to better understand the DNN performance.
%expose weaknesses in the current DNN development process, such as quality and distribution of data. 
%They can also help developers better understand DNN's performance and collect appropriate data for training and testing. 
%OLD:
%The confusion matrix computed for the inputs that pass the run-time check 
%(along with the estimated transition probabilities in the probabilistic abstraction $\alpha^{true}$) 
%is shown in Table~\ref{tab:trans_prob_vc_5}, Appendix~\ref{sec:tables}.
For instance, the DNN performs best for inputs lying on the center-line, which
%and degrades as the plane proceeds towards the right or the left. 
can be attributed to training being done mainly using scenarios where the plane follows the center-line. 
%Though the overall accuracy of the classifier (66.39\% for $\vcte$ and 69.99\% for $\vhe$) is not very high, 
%Note also that %for the $\vhe$ prediction, 
The model appears to perform better when the plane is heading left, as opposed to heading right, which may be due to camera position. These observations can be used by developers to improve the model, by training on more scenarios.
Note also that the model does not make `blatant' errors, mistaking inputs on the \emph{left} as being  on the \emph{right} (of center-line) or vice-versa (see e.g., entries with zero observations).   Formal proofs can provide guarantees of absence of such transitions.

\begin{figure*}[t]
    \centering
    \begin{subfigure}[t]{0.3\textwidth}
        \centering
        \includegraphics[width=\textwidth]{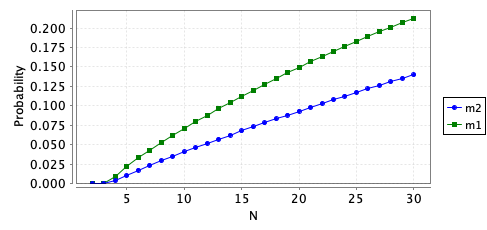}
        %\caption{The probability of $\vcte$ failure ($P=? [ F (\vcte=-1)]$)}
        \caption{\emph{Property 1}}
    \end{subfigure}
    ~
    \begin{subfigure}[t]{0.3\textwidth}
        \centering
        \includegraphics[width=\textwidth]{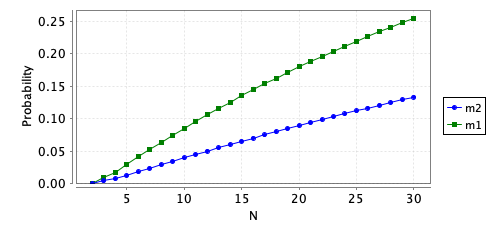}
        %\caption{The probability of $\vhe$ failure ($P=? [ F (\vhe=-1)]$)}
        \caption{\emph{Property 2}}
    \end{subfigure}%
    ~
    % \begin{subfigure}[t]{0.475\textwidth}
    %     \centering
    %     \includegraphics[height=2.0in]{taxinet_plots/taxinet_conditional.pdf}
    %     \caption{The probability of $\vcte$ or $\vhe$ failure given the input passes the run-time check ($P=? [ F (\vcte=-1 \vee \vhe=-1) \mid G(v=1)]$)}
    % \end{subfigure}%  
    %~
    \begin{subfigure}[t]{0.3\textwidth}
        \centering
        \includegraphics[width=\textwidth]{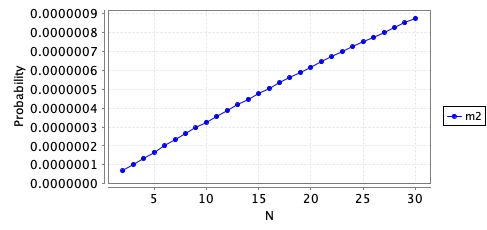}
        %\caption{The probability of abort due to run-time check failure ($P=? [ F (v=0~\&~i=M)]$)}
        \caption{\emph{Property 3}}
    \end{subfigure}% 
    \vspace{-0.3cm}
    \caption{Probabilistic model checking results via \prism{}}
    %(a) The probability of $\vcte$ failure ($P=? [ F (\vcte=-1)]$); (b) The probability of $\vhe$ failure ($P=? [ F (\vhe=-1)]$); (c) The probability of abort due to run-time check failure ($P=? [ F (v=0~\&~i=M)]$)} 
    \label{fig:prism_results_5bins}\vspace{-0.5cm}
\end{figure*}

\paragraph{Analysis.} We analyzed $m_1$ and $m_2$ with respect to the two PCTL properties, $P=? [ F (\vcte=-1)]$ ({\em Property 1}), and 
$P=? [ F (\vhe=-1)]$ ({\em Property 2}). The airplane is assumed to start from a initial position on the center-line and heading straight.  For $m_2$, i.e. the model with a run-time guard, 
%we implemented a fail-safe mechanism, as described in Section ~\ref{sec:abs_refine}, that repeatedly performs sensor readings until either the run-time check passes or the number of consecutive failures reaches a threshold and the system aborts. 
we also evaluate the probability of the TaxiNet system going to the abort state using the property, $P =?[F (v = 0~\&~i = M )$ ({\em Property 3}), where $M$ is the threshold for the number of consecutive run-time check failures. %For the purpose of our analysis, $M$ is fixed as 10. 

\begin{figure}[t]
    \centering
    \begin{subfigure}[b]{0.4\textwidth}
     \centering
     \includegraphics[width=\textwidth]{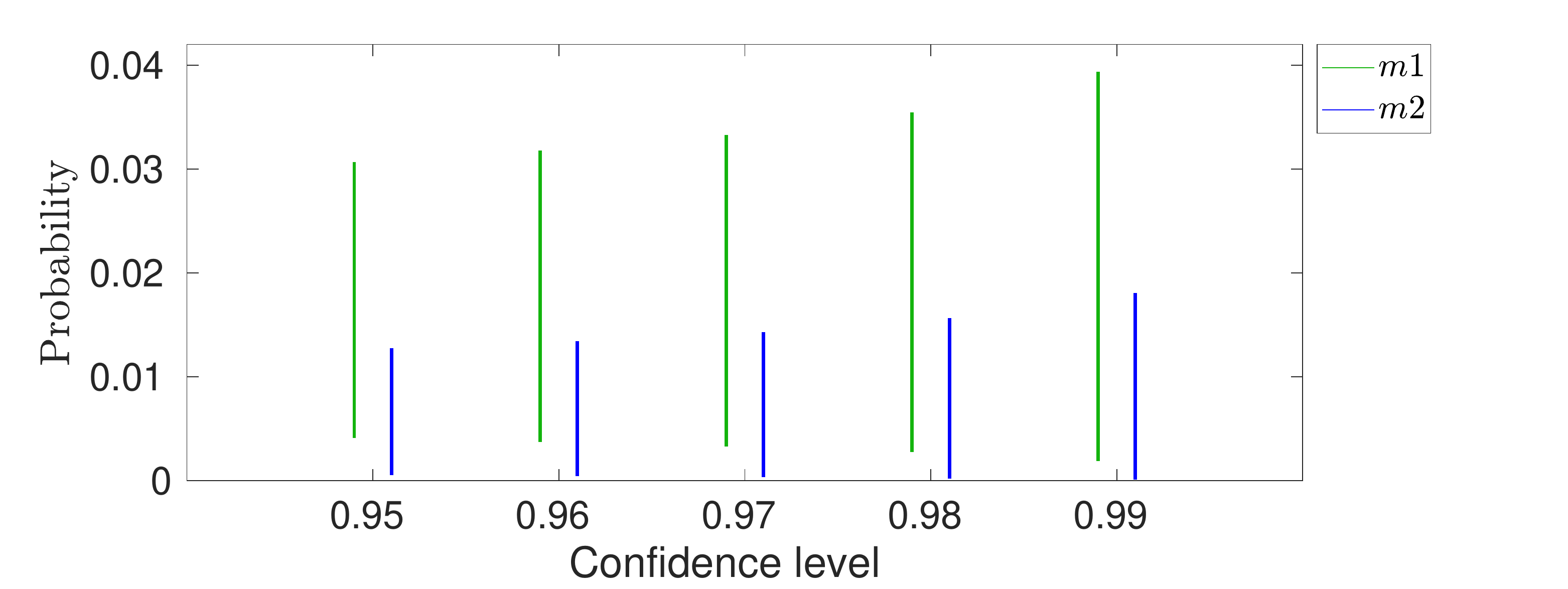}
        \caption{\emph{Property 1}}
        %: (a) Property $P=? [ F (\vcte=-1)]$; (b) Property $P=? [ F (\vhe=-1)]$ }
        \label{fig:CI_taxinet}
\end{subfigure}
~
    \begin{subfigure}[b]{0.4\textwidth}
        \centering
        \includegraphics[width=\textwidth]{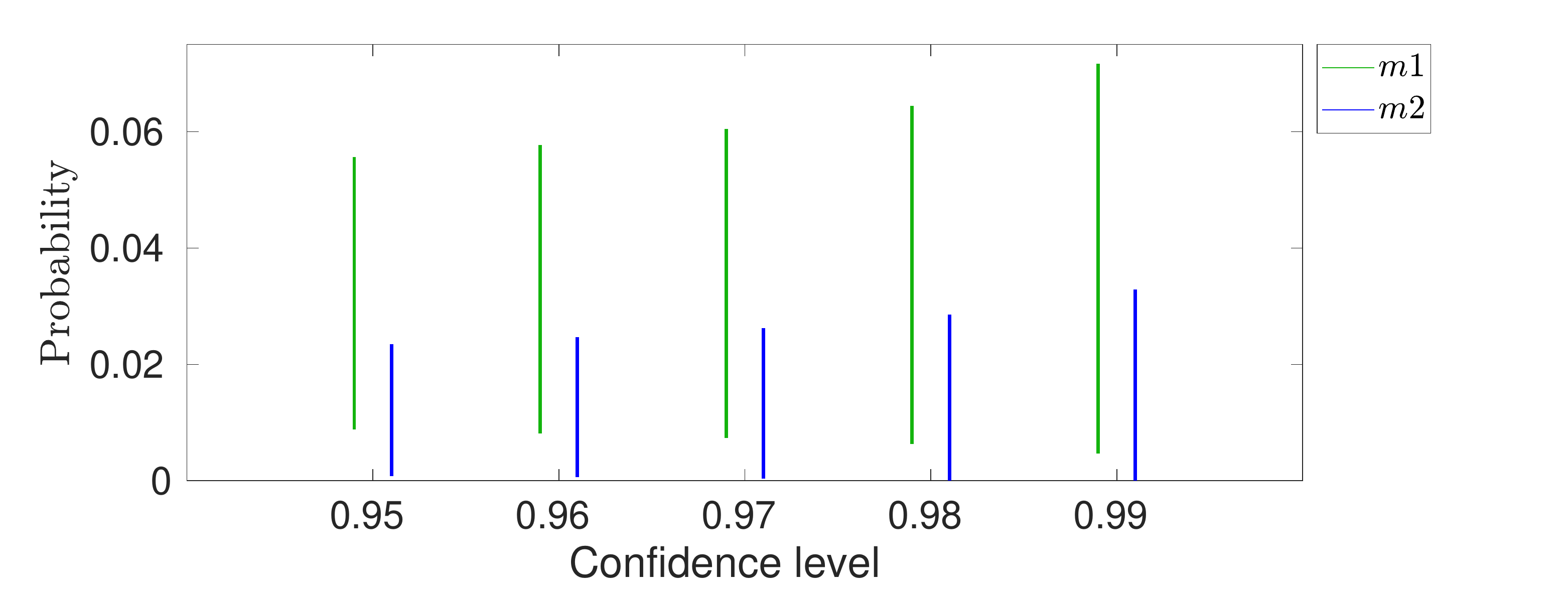}
        %\caption{Confidence intervals for the property $P=? [ F (\vcte=-1)]$}
        \caption{\emph{Property 2}}
        \label{fig:ci_vc_p1}
    \end{subfigure}
    % ~
    % \begin{subfigure}[b]{0.3\textwidth}
    %    \centering
    %    \includegraphics[width=\textwidth] {taxinet_plots/fact/5bins/taxinet_m2_prop3.pdf}
    %      % \caption{Confidence intervals for the property $P=? [ F (v=0)]$}
    %     \caption{\emph{Property 3}}
    %    \label{fig:ci_vc_p3}
    % \end{subfigure}
    \vspace{-0.3cm}
       \caption{Confidence interval results via \fact{}}
        % : (a) Property $P=? [ F (\vcte=-1)]$; (b) Property $P=? [ F (\vhe=-1)]$; (c) Property $P=? [ F (v=0~\wedge~i=M)$ }
       \label{fig:CI}
       \vspace{-0.6cm}
\end{figure}

The probabilities of these properties being satisfied, calculated by \prism{}, are shown in Figure~\ref{fig:prism_results_5bins}, where $N$ is a constant in the DTMCs that dictates the length of the finite-time horizon 
%(measured in terms of the number of airplane movement steps) 
considered for the analysis.
%Figure~\ref{fig:prism_results_5bins}c shows this probability as a function of $N$ ($M=10$). 
The confidence intervals computed with \fact{} are shown in Figure~\ref{fig:CI}, at different confidence levels (0.95 to 0.99), for $N=4$. For computing the intervals, we ignore the transitions in the DTMCs that were not observed in our data (see Appendix~\ref{sec:scale_fact} for more details).
%\corina{to be updated with new results}. We present the confidence intervals for the probabilities of satisfying the properties, at different confidence levels, for $N=4$. 

The \prism{} analysis 
scales well;
%can be done quite cheaply in terms of computational resources. 
e.g., evaluating \emph{Property 1} for model $m_2$ ($N=30$) requires less than 0.1 seconds on an M1 MacBook Pro, 16 GB RAM. The numbers are similar for other queries. 
%While the \prism{} analysis is highly scalable, we find 
However, the confidence analysis does 
%comprised of a parametric model checking stage followed by confidence interval inference stage, 
not scale as well; we could not go beyond $N=4$ for a timeout of two hours, with \emph{Property 1} hardest to check. 
%For instance, on a Windows OS machine with 11th Gen Intel(R) Core(TM) i7 processor and 64GB RAM,  \fact{}-based confidence analysis times out beyond $N=4$ for $m_1$ and $m_2$, assuming a 2 hour timeout.
Newer work, fPMC \cite{Fang22CGA}, addresses these scalability challenges but we found it not yet mature enough to be applied to our models.
%; we plan to improve the tool in future work.

%\corina{All the experiments were run on a machine XXX}
\paragraph{Discussion and Lessons Learned.}
%\ravi{I actually like this and the following two paras. corina: ok thanks :) In this section, we want to highlight 3 aspects of our work:(i) even with a complex DNN, we are actually able to perform a formal analysis, and in fact, do so quite cheaply in terms of computational resources, (ii) run-time guards can be helpful so system designers should think about them, (iii) the confidence analysis does not really scale} 
The experiments 
demonstrate the feasibility of our approach, which enables reasoning about a complex DNN interacting with conventional (discrete-time) components via a simple probabilistic abstraction. Our analysis not only provides qualitative (i.e., an error is reachable or not) but also quantitative (i.e., likelihood of error) results, helping developers assess the risk associated with the analyzed scenario.

%our indicate that the TaxiNet system is safe with less than 10\% probability of failure for paths shorten than 10 steps. However, this probability increases with the length of the taxiway reaching around 25\% for N=30. 

%\corina{how? this does not appear from the figures} The system is more vulnerable in turning more than a certain threshold (violating property 2), compared to going farther away from the center-line (violating property 1). This could be attributed to more number of discrete states and a more involved controller logic for $\vcte$.  It is interesting to note that when we consider the classification accuracy of the DNN model stand-alone, the accuracy of $\vcte$ prediction is worse than the $\vhe$ prediction. However, the closed-loop system analysis highlights that the mis-classification of $\vhe$ leads to the property violation more frequently than mis-classification wrt. $\vcte$.

The results 
%in Figure~\ref{fig:prism_results_5bins}(a,b) 
highlight the benefit of the run-time guards in improving the safety of the overall system; see Figures~\ref{fig:prism_results_5bins}(a,b) for lower error probabilities and Figures~\ref{fig:CI}(a,b) for tighter intervals for $m_2$. %Figures~\ref{fig:prism_results_5bins},\ref{fig:CI}(c) show that 
The probability of aborting %due to consecutive run-time check failures 
is very small, indicating the efficacy of the fail-safe mechanism (see Figures~\ref{fig:prism_results_5bins}(c),\ref{fig:ci_vc_p3}).
%The probability of system failure decreases from $m_1$ to $m_2$ with respect to both properties, specifically for \emph{Property 2} with around 48\% decrease at $N=30$. 
%There are two lessons learned from these experiments: (1)  {\em the run-time guard is effective} in lowering the error probabilities, and  (2) 
More importantly, since the DNN demonstrates higher accuracy on the inputs where the run-time check passes, the results also indicate that {\em improved accuracy of the DNN translates into improved safety}. 
%The results in Figure~\ref{fig:CI} display confidence intervals instead of point-wise probabilities. 
The computed probabilities and confidence intervals can be examined by developers and regulators to ensure that system safety is met at required levels. If the confidence intervals are too large, they can be made tighter by adding more data, as guided by the confusion matrices.

\section{Conclusion}
\label{sec:conclusion}
%\divya{added this based on Huafeng's feedback}
%We presented an approach that provides a solution to address system-level safety analysis of autonomous systems with imperfect machine-learning (ML) perception. The proposed abstraction helps separate the concerns of ML and conventional system development and evaluation. It also enables the integration of heterogeneous artifacts for consistent analysis, with the probabilistic approach and modeling methods. We demonstrated the method through a case study that involves a complex perception DNN working side-by-side with traditional components. The results from the study show not only the rate of potential requirement violations or improvement on the correctness, but also provide insights that can be used in quantitative safety assessment for AI/ML-enabled systems.  This is, potentially, an important step to fill one of the gaps for future AI certification. 

%\divya{moved this from conclusion}
We demonstrated a method for the analysis of the safety of autonomous systems that use  complex DNNs for visual perception. Our abstraction helps separate the concerns of DNN and conventional system development and evaluation. It also enables the integration of heterogeneous artifacts from DNN-specific analysis and system-level probabilistic model checking. The approach produces not only qualitative results but also
also provides insights that can be used in quantitative safety assessment for AI/DNN-enabled systems.  
%This information could be potentially used as proofs for the compliance with safety objectives defined in EASA’s AI guidance. This research could  help fill one of the gaps of quantitative evaluation in future AI certification
This is, potentially, an important step to fill one of the gaps of quantitative evaluation for future AI certification~\cite{easa}. 

%We demonstrated the method through a case study that involves a complex perception DNN working side-by-side with traditional components. 
Future work involves more experimentation with image data sets representing a wide variety of environment conditions. We also plan to refine our models, inducing finer partitions on the DNN, and validate them through hardware simulations. Finally, we plan to study the composition of safety proofs for the system analyzed in different scenarios.

\newpage
%% Acknowledgments
%\paragraph{\textbf{Acknowledgments.}}

%% Bibliography
%\bibliographystyle{splncs04}
%\usepackage[numbers,sort&compress]{natbib}
%\bibliographystyle{splncs04nat}
%\bibliography{bibfile}

\newpage
%% Appendix
\appendix
\section{Confusion Matrices and Transition Probabilities}
\label{sec:tables}

\begin{table*}
\resizebox{\textwidth}{!}{
\begin{tabular}
{|c|c|c|c|c|c|}\hline
{\textbf{Accuracy}}&\textbf{Grd Truth Label}&\textbf{Tot \# inputs}&\textbf{Actual Label}&\textbf{\# inputs}&\textbf{Tr Prob} \\\hline
\multirow{25}{*}{\textbf{$\vcte$ (62.39\%)}}&\multirow{5}{*}{3}&\multirow{5}{*}{398}&3&101&0.254\\\cline{4-6}
&&&1&145&0.364\\\cline{4-6}
&&&0&152&0.382\\\cline{4-6}
&&&2&0&0.0\\\cline{4-6}
&&&4&0&0.0\\\cline{2-6}

&\multirow{5}{*}{1}&\multirow{5}{*}{2693}&3&11&0.004\\\cline{4-6}
&&&1&992&0.368\\\cline{4-6}
&&&0&1686&0.627\\\cline{4-6}
&&&2&4&0.001\\\cline{4-6}
&&&4&0&0.0\\\cline{2-6}

&\multirow{5}{*}{0}&\multirow{5}{*}{4084}&3&0&0.0\\\cline{4-6}
&&&1&40&0.01\\\cline{4-6}
&&&0&3440&0.842\\\cline{4-6}
&&&2&604&0.148\\\cline{4-6}
&&&4&0&0.0\\\cline{2-6}

&\multirow{5}{*}{2}&\multirow{5}{*}{3304}&3&0&0.0\\\cline{4-6}
&&&1&4&0.001\\\cline{4-6}
&&&0&1119&0.339\\\cline{4-6}
&&&2&2173&0.658\\\cline{4-6}
&&&4&8&0.002\\\cline{2-6}

&\multirow{5}{*}{4}&\multirow{5}{*}{629}&3&0&0.0\\\cline{4-6}
&&&1&0&0.0\\\cline{4-6}
&&&0&151&0.24\\\cline{4-6}
&&&2&254&0.404\\\cline{4-6}
&&&4&224&0.356\\\hline

\multirow{9}{*}{\textbf{$\vhe$ (69.99\%)}}&\multirow{3}{*}{1}&\multirow{3}{*}{2101}&1&2010&0.957\\\cline{4-6}
&&&0&91&0.043\\\cline{4-6}
&&&2&0&0.0\\\cline{2-6}
&\multirow{3}{*}{0}&\multirow{3}{*}{7035}&1&2139&0.304\\\cline{4-6}
&&&0&4748&0.675\\\cline{4-6}
&&&2&148&0.021\\\cline{2-6}
&\multirow{3}{*}{2}&\multirow{3}{*}{1972}&1&211&0.107\\\cline{4-6}
&&&0&744&0.377\\\cline{4-6}
&&&2&1017&0.516\\\hline
\end{tabular}}
\caption{Results computed on TaxiNet for 11108 inputs.
%for all inputs. Total of 11108 inputs. The regression model MAE for $\vcte$ is 1.185 and for $\vhe$ is 7.860.
}
%\end{scriptsize}
\label{tab:trans_prob_5}
\end{table*}
\begin{table*}
\resizebox{\textwidth}{!}{
%\begin{scriptsize}
\begin{tabular}
{|c|c|c|c|c|c|}\hline
{\textbf{Accuracy}}&\textbf{Grd Truth Label}&\textbf{Tot \# inputs}&\textbf{Actual Label}&\textbf{\# inputs}&\textbf{Tr Prob} \\\hline
\multirow{25}{*}{\textbf{$\vcte$ (65.78\%)}}&\multirow{5}{*}{3}&\multirow{5}{*}{330}&3&100&0.303\\\cline{4-6}
&&&1&126&0.382\\\cline{4-6}
&&&0&104&0.315\\\cline{4-6}
&&&2&0&0.0\\\cline{4-6}
&&&4&0&0.0\\\cline{2-6}

&\multirow{5}{*}{1}&\multirow{5}{*}{2168}&3&11&0.005\\\cline{4-6}
&&&1&951&0.439\\\cline{4-6}
&&&0&1206&0.556\\\cline{4-6}
&&&2&0&0.0\\\cline{4-6}
&&&4&0&0.0\\\cline{2-6}

&\multirow{5}{*}{0}&\multirow{5}{*}{3185}&3&0&0.0\\\cline{4-6}
&&&1&32&0.01\\\cline{4-6}
&&&0&2984&0.909\\\cline{4-6}
&&&2&259&0.081\\\cline{4-6}
&&&4&0&0.0\\\cline{2-6}

&\multirow{5}{*}{2}&\multirow{5}{*}{2897}&3&0&0.0\\\cline{4-6}
&&&1&2&0.001\\\cline{4-6}
&&&0&1069&0.369\\\cline{4-6}
&&&2&1820&0.628\\\cline{4-6}
&&&4&6&0.002\\\cline{2-6}

&\multirow{5}{*}{4}&\multirow{5}{*}{514}&3&0&0.0\\\cline{4-6}
&&&1&0&0.0\\\cline{4-6}
&&&0&119&0.232\\\cline{4-6}
&&&2&178&0.346\\\cline{4-6}
&&&4&217&0.422\\\hline

\multirow{9}{*}{\textbf{$\vhe$ (76\%)}}&\multirow{3}{*}{1}&\multirow{3}{*}{1991}&1&1902&0.955\\\cline{4-6}
&&&0&89&0.045\\\cline{4-6}
&&&2&0&0.0\\\cline{2-6}
&\multirow{3}{*}{0}&\multirow{3}{*}{5538}&1&1388&0.25\\\cline{4-6}
&&&0&4025&0.727\\\cline{4-6}
&&&2&125&0.023\\\cline{2-6}
&\multirow{3}{*}{2}&\multirow{3}{*}{1596}&1&76&0.0.48\\\cline{4-6}
&&&0&512&0.32\\\cline{4-6}
&&&2&1008&0.632\\\hline
\end{tabular}}
\caption{Results for TaxiNet, on inputs that pass the DNN check (9125 out of 11108). 
%Out of 11108 inputs, 9125 inputs pass the check. For the remaining inputs, the regression model MAE for $\vcte$ is 1.078 and for $\vhe$ is 6.393.
}
\label{tab:trans_prob_vc_5}
\end{table*}

\newpage

\section{Non-parametric DTMC models for \prism{}}
\label{sec:dtmc_prism}

\subsection{Model $m_1$ (no run-time guard; $v=1$ always)}
\label{sec:dtmc_prism_no_vc}

%\lstinputlisting[language={Prism}, numbers=left, rulesepcolor=\color{black}, rulecolor=\color{black}, breaklines=true, breakatwhitespace=true, firstnumber=1, firstline=1, caption={3 bin DTMC model without run-time guards}, label={dtmc_prism3_no_vc}] {prism_code/taxinet_Boeing3bins.pm}

\begin{lstlisting}[language={Prism}, numbers=left, rulesepcolor=\color{black}, rulecolor=\color{black}, breaklines=true, breakatwhitespace=true, firstnumber=1, firstline=1, 
caption={DTMC model without run-time guards},
label={dtmc_prism5_no_vc_small}]
dtmc

const N;

module taxinet

	cte : [-1..4] init 0;//0:center, 1,3:left, 2,4:right, -1:error
	he: [-1..2] init 0;//0:no angle, 1:negative, 2:positive, -1:error
	v: [0..1] init 1;//1:certified, 0:not certified
	cte_est: [0..4] init 0;
	he_est: [0..2] init 0; 
	a: [0 .. 2] init 0;//0:go straight, 1:turn left, 2: turn right
	pc:[0..5] init 0; // program counter
	k:[1..N] init 1; // steps

	// run-time guard: assume we can certify all inputs	
	[]  pc=0 -> 1: (v'=1) & (pc'=1);	

	// NN imperfect perception 
	[] cte=0 & v=1 & pc=1 -> 0.842: (cte_est'=0) & (pc'=2) + 
				 0.01: (cte_est'=1) & (pc'=2) + 
				 0.148: (cte_est'=2) & (pc'=2) + 
				 0.0: (cte_est'=3) & (pc'=2) + 
				 0.0: (cte_est'=4) & (pc'=2);
	[] cte=1 & v=1 & pc=1 -> 0.627: (cte_est'=0) & (pc'=2) + 
				 0.368: (cte_est'=1) & (pc'=2) + 
				 0.001: (cte_est'=2) & (pc'=2) + 
				 0.004: (cte_est'=3) & (pc'=2) + 
				 0.0: (cte_est'=4) & (pc'=2);
	[] cte=2 & v=1 & pc=1 -> 0.339: (cte_est'=0) & (pc'=2) + 
				 0.001: (cte_est'=1) & (pc'=2) + 
				 0.658: (cte_est'=2) & (pc'=2) + 
				 0.0: (cte_est'=3) & (pc'=2) + 
				 0.002: (cte_est'=4) & (pc'=2);
	[] cte=3 & v=1 & pc=1 -> 0.382: (cte_est'=0) & (pc'=2) + 
				 0.364: (cte_est'=1) & (pc'=2) + 
				 0.0: (cte_est'=2) & (pc'=2) + 
				 0.254: (cte_est'=3) & (pc'=2) + 
				 0.0: (cte_est'=4) & (pc'=2);
	[] cte=4 & v=1 & pc=1 -> 0.24: (cte_est'=0) & (pc'=2) + 
				 0.0: (cte_est'=1) & (pc'=2) + 
				 0.404: (cte_est'=2) & (pc'=2) + 
				 0.0: (cte_est'=3) & (pc'=2) + 
				 0.356: (cte_est'=4) & (pc'=2);


	[] he=0 & v=1 & pc=2 -> 0.675: (he_est'=0) & (pc'=3) + 
				0.304: (he_est'=1) & (pc'=3) + 
				0.021: (he_est'=2) & (pc'=3); 	
	[] he=1 & v=1 & pc=2 -> 0.043: (he_est'=0) & (pc'=3) + 
				0.957: (he_est'=1) & (pc'=3) + 
				0: (he_est'=2) & (pc'=3);
	[] he=2 & v=1 & pc=2 -> 0.377: (he_est'=0) & (pc'=3) + 
				0.107: (he_est'=1) & (pc'=3) + 
				0.516: (he_est'=2) & (pc'=3);
	
	//controller: 
	[] cte_est=0 & he_est=0 & pc=3 -> 1 : (a'=0) & (pc'=4);
	[] cte_est=0 & he_est=1 & pc=3 -> 1 : (a'=2) & (pc'=4);
	[] cte_est=0 & he_est=2 & pc=3 -> 1 : (a'=1) & (pc'=4);
	
	[] (cte_est=1 | cte_est=3) & he_est=0 & pc=3 -> 1 : (a'=2) & (pc'=4);
	[] (cte_est=1 | cte_est=3) & he_est=1 & pc=3 -> 1 : (a'=2) & (pc'=4);
	[] (cte_est=1 | cte_est=3) & he_est=2 & pc=3 -> 1 : (a'=0) & (pc'=4);
	
	[] (cte_est=2 | cte_est=4) & he_est=0 & pc=3 -> 1 : (a'=1) & (pc'=4);
	[] (cte_est=2 | cte_est=4) & he_est=1 & pc=3 -> 1 : (a'=0) & (pc'=4);
	[] (cte_est=2 | cte_est=4) & he_est=2 & pc=3 -> 1 : (a'=1) & (pc'=4);

	// airplane dynamics: 
	[] he=1 & a=1 & pc=4 & k<N -> 1 : (he'=-1) & (pc'=5); // error
	[] he=2 & a=2 & pc=4 & k<N -> 1 : (he'=-1) & (pc'=5); // error

	[] cte=0 & he=0 & a=0 & pc=4 & k<N -> 1 : (pc'=0) & (k'=k+1);
	[] cte=0 & he=0 & a=1 & pc=4 & k<N -> 1 : (cte'=1) & (he'=1) & (pc'=0) & (k'=k+1);
	[] cte=0 & he=0 & a=2 & pc=4 & k<N -> 1 : (cte'=2) & (he'=2) & (pc'=0) & (k'=k+1);

	[] cte=0 & he=1 & a=0 & pc=4 & k<N -> 1 : (cte'=1) & (pc'=0) & (k'=k+1);
	[] cte=0 & he=1 & a=2 & pc=4 & k<N -> 1 : (he'=0) & (pc'=0) & (k'=k+1);

	[] cte=0 & he=2 & a=0 & pc=4 & k<N -> 1 : (cte'=2) & (pc'=0) & (k'=k+1);
	[] cte=0 & he=2 & a=1 & pc=4 & k<N -> 1 : (he'=0) & (pc'=0) & (k'=k+1);

	// left-side dynamics:
	[] cte=1 & he=0 & a=0 & pc=4 & k<N -> 1 : (pc'=0) & (k'=k+1);
	[] cte=1 & he=0 & a=1 & pc=4 & k<N -> 1 : (cte'=3) & (he'=1) & (pc'=0) & (k'=k+1);
	[] cte=1 & he=0 & a=2 & pc=4 & k<N -> 1 : (cte'=0) & (he'=2) & (pc'=0) & (k'=k+1);

	[] cte=1 & he=1 & a=0 & pc=4 & k<N -> 1 : (cte'=3) & (pc'=0) & (k'=k+1);
	[] cte=1 & he=1 & a=2 & pc=4 & k<N -> 1 : (he'=0) & (pc'=0) & (k'=k+1);

	[] cte=1 & he=2 & a=0 & pc=4 & k<N -> 1 : (cte'=0) & (pc'=0) & (k'=k+1);
	[] cte=1 & he=2 & a=1 & pc=4 & k<N -> 1 : (he'=0) & (pc'=0) & (k'=k+1);

	[] cte=3 & he=0 & a=0 & pc=4 & k<N -> 1 : (pc'=0) & (k'=k+1);
	[] cte=3 & he=0 & a=1 & pc=4 & k<N -> 1 : (cte'=-1) & (pc'=5); //error
	[] cte=3 & he=0 & a=2 & pc=4 & k<N -> 1 : (cte'=1) & (he'=2) & (pc'=0) & (k'=k+1);

	[] cte=3 & he=1 & a=0 & pc=4 & k<N -> 1 : (cte'=-1) & (pc'=5); //error
	[] cte=3 & he=1 & a=2 & pc=4 & k<N -> 1 : (he'=0) & (pc'=0) & (k'=k+1);

	[] cte=3 & he=2 & a=0 & pc=4 & k<N -> 1 : (cte'=1) & (pc'=0) & (k'=k+1);
	[] cte=3 & he=2 & a=1 & pc=4 & k<N -> 1 : (he'=0) & (pc'=0) & (k'=k+1);

	// right-side dynamics:
	[] cte=2 & he=0 & a=0 & pc=4 & k<N -> 1 : (pc'=0) & (k'=k+1);
	[] cte=2 & he=0 & a=1 & pc=4 & k<N -> 1 : (cte'=0) & (he'=1) & (pc'=0) & (k'=k+1);
	[] cte=2 & he=0 & a=2 & pc=4 & k<N -> 1 : (cte'=4) & (he'=2) & (pc'=0) & (k'=k+1);

	[] cte=2 & he=1 & a=0 & pc=4 & k<N -> 1 : (cte'=0) & (pc'=0) & (k'=k+1);
	[] cte=2 & he=1 & a=2 & pc=4 & k<N -> 1 : (he'=0) & (pc'=0) & (k'=k+1);

	[] cte=2 & he=2 & a=0 & pc=4 & k<N -> 1 : (cte'=4) & (pc'=0) & (k'=k+1);
	[] cte=2 & he=2 & a=1 & pc=4 & k<N -> 1 : (he'=0) & (pc'=0) & (k'=k+1);

	[] cte=4 & he=0 & a=0 & pc=4 & k<N -> 1 : (pc'=0) & (k'=k+1);
	[] cte=4 & he=0 & a=1 & pc=4 & k<N -> 1 : (cte'=2) & (he'=1) & (pc'=0) & (k'=k+1);
	[] cte=4 & he=0 & a=2 & pc=4 & k<N -> 1 : (cte'=-1) & (pc'=5); //error

	[] cte=4 & he=1 & a=0 & pc=4 & k<N -> 1 : (cte'=2) & (pc'=0) & (k'=k+1);
	[] cte=4 & he=1 & a=2 & pc=4 & k<N -> 1 : (he'=0) & (pc'=0) & (k'=k+1);

	[] cte=4 & he=2 & a=0 & pc=4 & k<N -> 1 : (cte'=-1) & (pc'=5); //error
	[] cte=4 & he=2 & a=1 & pc=4 & k<N -> 1 : (he'=0) & (pc'=0) & (k'=k+1);

endmodule
\end{lstlisting} 

\subsection{Model $m_2$ (with run-time guard)}
\label{sec:dtmc_prism_vc}

%\lstinputlisting[language={Prism}, numbers=left, rulesepcolor=\color{black}, rulecolor=\color{black}, breaklines=true, breakatwhitespace=true, firstnumber=1, firstline=1,
%caption={3 bin DTMC model with run-time guards},
%label={dtmc_prism3_vc}]{prism_code/taxinet_Boeing3bins_vc.pm}

\begin{lstlisting}[language={Prism}, numbers=left, rulesepcolor=\color{black}, rulecolor=\color{black}, breaklines=true, breakatwhitespace=true, firstnumber=1, firstline=1, 
caption={DTMC model with run-time guards},
label={dtmc_prism5_vc}]
dtmc

const N;
const M=10;

module taxinet

	cte : [-1..4] init 0;
	he: [-1..2] init 0;
	v: [0..1] init 1;
	cte_est: [0..4] init 0;
	he_est: [0..2] init 0; 
	a: [0 .. 2] init 0; 
	pc:[0..5] init 0; 
	k:[1..N] init 1;
	
	//number of sensor readings
	i:[0..M] init 0;

	// run-time guard based on Prophecy rules
	[]  pc=0 & i<M -> 0.821: (v'=1) & (pc'=1) & (i'=0) + 
				0.179: (v'=0) & (i'=i+1);//repeat reading

	// NN imperfect perception 
	[] cte=0 & v=1 & pc=1 -> 0.909: (cte_est'=0) & (pc'=2) + 
				 0.01: (cte_est'=1) & (pc'=2) + 
				 0.081: (cte_est'=2) & (pc'=2) + 
				 0.0: (cte_est'=3) & (pc'=2) + 
				 0.0: (cte_est'=4) & (pc'=2);
	[] cte=1 & v=1 & pc=1 -> 0.556: (cte_est'=0) & (pc'=2) + 
				 0.439: (cte_est'=1) & (pc'=2) + 
				 0.0: (cte_est'=2) & (pc'=2) + 
				 0.005: (cte_est'=3) & (pc'=2) + 
				 0.0: (cte_est'=4) & (pc'=2);
	[] cte=2 & v=1 & pc=1 -> 0.369: (cte_est'=0) & (pc'=2) + 
				 0.001: (cte_est'=1) & (pc'=2) + 
				 0.628: (cte_est'=2) & (pc'=2) + 
				 0.0: (cte_est'=3) & (pc'=2) + 
				 0.002: (cte_est'=4) & (pc'=2);
	[] cte=3 & v=1 & pc=1 -> 0.315: (cte_est'=0) & (pc'=2) + 
				 0.382: (cte_est'=1) & (pc'=2) + 
				 0.0: (cte_est'=2) & (pc'=2) + 
				 0.303: (cte_est'=3) & (pc'=2) + 
				 0.0: (cte_est'=4) & (pc'=2);
	[] cte=4 & v=1 & pc=1 -> 0.232: (cte_est'=0) & (pc'=2) + 
				 0.0: (cte_est'=1) & (pc'=2) + 
				 0.346: (cte_est'=2) & (pc'=2) + 
				 0.0: (cte_est'=3) & (pc'=2) + 
				 0.422: (cte_est'=4) & (pc'=2);



	[] he=0 & v=1 & pc=2 -> 0.727: (he_est'=0) & (pc'=3) + 
				0.25: (he_est'=1) & (pc'=3) + 
				0.023: (he_est'=2) & (pc'=3); 	
	[] he=1 & v=1 & pc=2 -> 0.045: (he_est'=0) & (pc'=3) + 
				0.955: (he_est'=1) & (pc'=3) + 
				0: (he_est'=2) & (pc'=3);
	[] he=2 & v=1 & pc=2 -> 0.32: (he_est'=0) & (pc'=3) + 
				0.048: (he_est'=1) & (pc'=3) + 
				0.632: (he_est'=2) & (pc'=3);
	
	// controller and airplane dynamics: same as before
	...
endmodule
\end{lstlisting}

\section{Parametric DTMC models for \fact{}}
\label{sec:dtmc_fact}

\subsection{Model $m_1$ (no run-time guard; $v=1$ always)}
\label{sec:dtmc_fact_no_vc}
\begin{lstlisting}[language={Prism}, numbers=left, rulesepcolor=\color{black}, rulecolor=\color{black}, breaklines=true, breakatwhitespace=true, firstnumber=1, firstline=1, 
caption={Parametric DTMC model without run-time guards},
label={dtmc_prism5_fact}]
dtmc

const N=4;
param double p = 3440 604 40;
param double q = 1686 992 4 11;
param double r = 1119 4 2173 8;
param double t = 152 145 101;
param double h = 151 254 224;

param double x = 4748 2139 148;
param double y = 91 2010;
param double z = 744 211 1017;

module taxinet

	cte : [-1..4] init 0;
	he: [-1..2] init 0;
	v: [0..1] init 1;
	cte_est: [0..4] init 0;
	he_est: [0..2] init 0; 
	a: [0 .. 2] init 0; 
	pc:[0..5] init 0; 
	k:[1..N] init 1;
	
	[] pc=0 -> 1: (v'=1)& (pc'=1);

	[] cte=0 & v=1 & pc=1 -> p1: (cte_est'=0) & (pc'=2) + 
				 p2: (cte_est'=1) & (pc'=2) + 
				 (1-p1-p2): (cte_est'=2) & (pc'=2);
	[] cte=1 & v=1 & pc=1 -> q1: (cte_est'=0) & (pc'=2) + 
				 q2: (cte_est'=1) & (pc'=2) + 
				 q3: (cte_est'=2) & (pc'=2) + 
				 (1-q1-q2-q3): (cte_est'=3) & (pc'=2);
	[] cte=2 & v=1 & pc=1 -> r1: (cte_est'=0) & (pc'=2) + 
				 r2: (cte_est'=1) & (pc'=2) + 
				 r3: (cte_est'=2) & (pc'=2) + 
				 (1-r1-r2-r3): (cte_est'=4) & (pc'=2);
	[] cte=3 & v=1 & pc=1 -> t1: (cte_est'=0) & (pc'=2) + 
				 t2: (cte_est'=1) & (pc'=2) + 
				 (1-t1-t2): (cte_est'=3) & (pc'=2); 
	[] cte=4 & v=1 & pc=1 -> h1: (cte_est'=0) & (pc'=2) + 
				 h2: (cte_est'=2) & (pc'=2) + 
				 (1-h1-h2): (cte_est'=4) & (pc'=2);


	[] he=0 & v=1 & pc=2 -> x1: (he_est'=0) & (pc'=3) + 
				x2: (he_est'=1) & (pc'=3) + 
				(1-x1-x2): (he_est'=2) & (pc'=3); 	
	[] he=1 & v=1 & pc=2 -> y1: (he_est'=0) & (pc'=3) + 
				(1-y1): (he_est'=1) & (pc'=3); 
	[] he=2 & v=1 & pc=2 -> z1: (he_est'=0) & (pc'=3) + 
				z2: (he_est'=1) & (pc'=3) + 
				(1-z1-z2): (he_est'=2) & (pc'=3);

	// controller and airplane dynamics: same as before
        ...

endmodule
\end{lstlisting}

\subsection{Model $m_2$ (with run-time guard)}
\label{sec:dtmc_fact_vc}
\begin{lstlisting}[language={Prism}, numbers=left, rulesepcolor=\color{black}, rulecolor=\color{black}, breaklines=true, breakatwhitespace=true, firstnumber=1, firstline=1, 
caption={Parametric DTMC model with run-time guards},
label={dtmc_prism5_vc_fact}]
dtmc

const N=4;
const M=10;

param double c = 7496 1629;

param double p = 2984 32 259;
param double q = 1206 951 11;
param double r = 1069 2 1820 6;
param double t = 104 126 100;
param double h = 119 178 217;

param double x = 4025 1388 125;
param double y = 89 1902;
param double z = 512 76 1008;

module taxinet

	cte : [-1..4] init 0;
	he: [-1..2] init 0;
	v: [0..1] init 1;
	cte_est: [0..4] init 0;
	he_est: [0..2] init 0; 
	a: [0 .. 2] init 0; 
	pc:[0..5] init 0; 
	k:[1..N] init 1;
	i:[0..M] init 0;

	[]  pc=0 & i<M -> c1: (v'=1) & (pc'=1) & (i'=0) + 
			  (1-c1): (v'=0) & (i'=i+1);//repeat reading

	[] cte=0 & v=1 & pc=1 -> p1: (cte_est'=0) & (pc'=2) + 
				 p2: (cte_est'=1) & (pc'=2) + 
				 (1-p1-p2): (cte_est'=2) & (pc'=2);
	[] cte=1 & v=1 & pc=1 -> q1: (cte_est'=0) & (pc'=2) + 
				 q2: (cte_est'=1) & (pc'=2) + 
				 (1-q1-q2): (cte_est'=3) & (pc'=2); 
	[] cte=2 & v=1 & pc=1 -> r1: (cte_est'=0) & (pc'=2) + 
				 r2: (cte_est'=1) & (pc'=2) + 
				 r3: (cte_est'=2) & (pc'=2) + 
				 (1-r1-r2-r3): (cte_est'=4) & (pc'=2);
	[] cte=3 & v=1 & pc=1 -> t1: (cte_est'=0) & (pc'=2) + 
				 t2: (cte_est'=1) & (pc'=2) + 
				 (1-t1-t2): (cte_est'=3) & (pc'=2); 
	[] cte=4 & v=1 & pc=1 -> h1: (cte_est'=0) & (pc'=2) + 
				 h2: (cte_est'=2) & (pc'=2) + 
				 (1-h1-h2): (cte_est'=4) & (pc'=2);


	[] he=0 & v=1 & pc=2 -> x1: (he_est'=0) & (pc'=3) + 
				x2: (he_est'=1) & (pc'=3) + 
				(1-x1-x2): (he_est'=2) & (pc'=3); 	
	[] he=1 & v=1 & pc=2 -> y1: (he_est'=0) & (pc'=3) + 
				(1-y1): (he_est'=1) & (pc'=3); 
	[] he=2 & v=1 & pc=2 -> z1: (he_est'=0) & (pc'=3) + 
				z2: (he_est'=1) & (pc'=3) + 
				(1-z1-z2): (he_est'=2) & (pc'=3);

	// controller and airplane dynamics: same as before
        ...	

endmodule
\end{lstlisting}

\section{Rules Mined from TaxiNet as Run-time Guards}
\label{sec:rules}
In previous work~\cite{KadronGPY21} we explored the use of rules mined from the TaxiNet model for the purpose of {\em explaining} correct and incorrect model behavior. The rules are in the form of $Pre \implies Post$, where the $Pre$ condition is in terms of neuron constraints in the latent space of the network and characterizes inputs on which the $Post$ condition (property of the network outputs) is satisfied. The following correctness property of the regression outputs was considered in~\cite{KadronGPY21}, $|\vcte^{*}-\vcte| \leq 1.0~\mathit{meters} \; \wedge \; |\vhe^{*}-\vhe| \leq 5~ \mathit{degrees}$, and rules for the satisfaction and violation of this property were extracted. We used the Prophecy tool~\cite{ASE19} to extract these rules. Given a set of inputs, the tool builds a set comprising of the neuron values corresponding to each input and a label of whether the $Post$ was satisfied or not, which in turn is fed to decision-tree learning. There could be more than 1 rules per label, each of which has 100\% precision on the input set and a metric of support representing the number of inputs satisfying the neuron constraints in the corresponding pre-condition.  

Leveraging this work, we extracted rules for valid and invalid behavior of the Taxinet regression model using a separate set of 5554 test inputs. The rules were extracted using the neurons from the three dense layers of the network (dense\_1 layer with 100, dense\_2 with 50 and dense\_3 with 10 neurons respectively).
Example of a rule for invalid behavior: 
\begin{multline*}
N_{1,85} <= -0.998~\wedge~N_{2,50} <= 3.31~\wedge~N_{1,84} <= -0.994~\wedge~N_{1,15} > -0.999 \\ 
~\wedge~N_{1,21} <= 1.711~\wedge~N_{1,70} <= 11.088~\wedge~N_{1,51} > -0.999~\wedge~N_{1,21} > -0.637 \implies \\
|\vcte^{*}-\vcte| > 1.0~\mathit{meters}~\lor~ |\vhe^{*}-\vhe| > 5~\mathit{degrees}
\end{multline*}

Here $N_{i,j}$ indicates the $j^{th}$ neuron in the $i^{th}$ dense layer. This rule has a support of 755, indicating that 755 (out of 5554) inputs satisfied the neuron constraints specified in $Pre$  and the DNN violated the correctness property for each of them.% The validation precision and recall of the rule is 97.79\% and 2\% respectively. 

For our case study, we experimented with deploying the rules for invalid behavior as run-time guards to catch inputs potentially leading to misbehavior. In order to obtain good coverage while maintaining high precision, 
we selected rules with the support greater than a threshold. When running on the dataset with 11108 inputs, inputs that satisfied any of the selected rules were discarded (1983 inputs), thereby abstaining the model from potential misbehavior. The regression model behavior improved (MAE for $\vcte$: 1.07, MAE for $\vhe$: 6.39), which in turn improved the accuracies on the classification abstraction as well ($\vcte$: 65.78\%, $\vhe$:76\%). 
The new confusion matrix (computed for the inputs that pass the run-time guard) and transition probabilities are shown in Table~\ref{tab:trans_prob_vc_5}. While we use rules for invalid behavior for the purpose of this demonstration, we could also employ other run-time guards using e.g., the rules for valid behavior, or rules for correct/incorrect behavior wrt. the discretized outputs and easily integrate them in our approach. 

\section{Limitations of \fact{}}
\label{sec:scale_fact}
\begin{figure}
\centering
\begin{minipage}{.45\textwidth}
  \centering
\includegraphics[width=\textwidth]{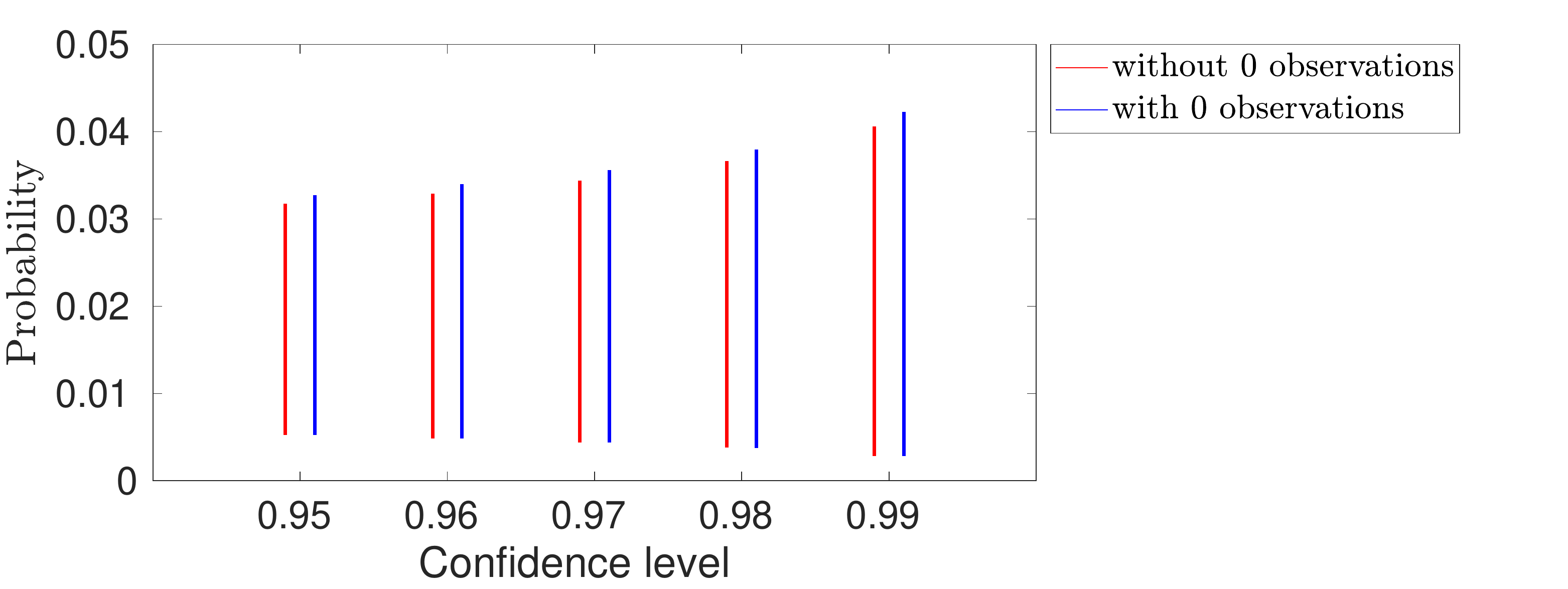}
        \captionof{figure}{Confidence interval results for $m_1$, \emph{Property 2}, $N=3$.}
        %: (a) Property $P=? [ F (\vcte=-1)]$; (b) Property $P=? [ F (\vhe=-1)]$ }
        \label{fig:CI_0obs}
\end{minipage}%
~
\begin{minipage}{.45\textwidth}
  \centering
\includegraphics[width=\textwidth] {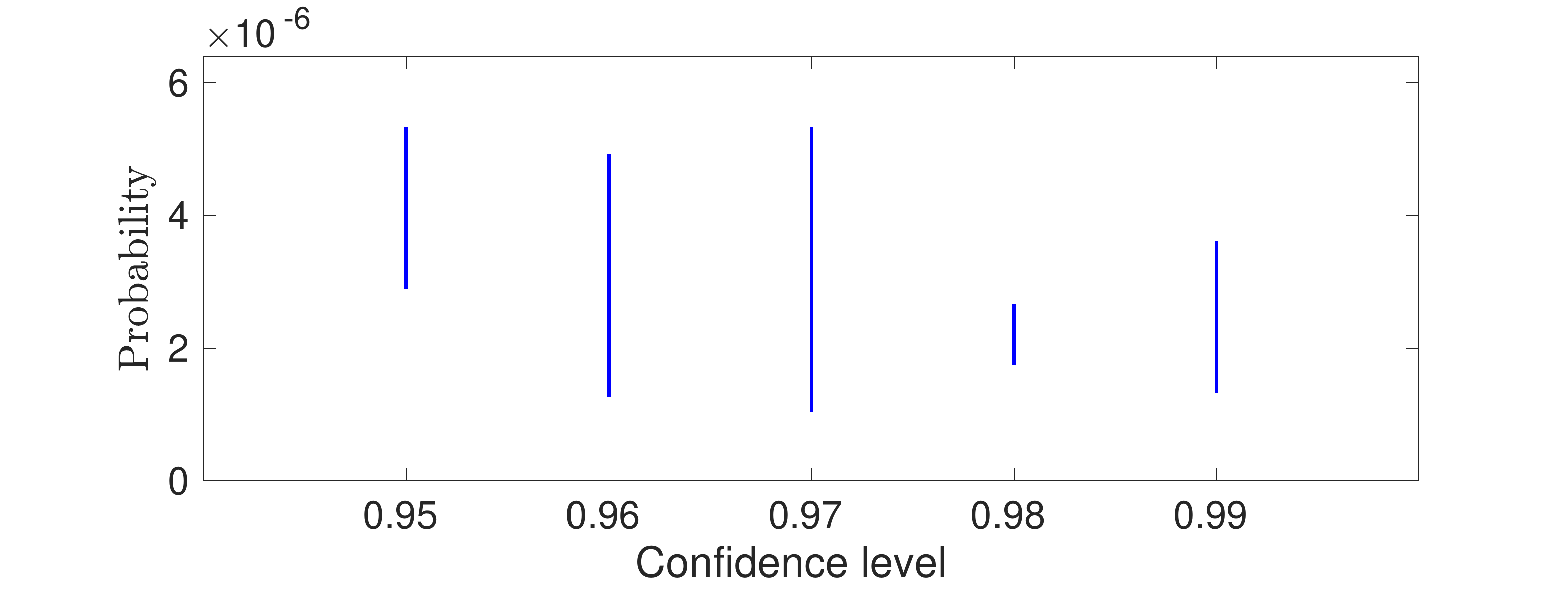}
    \captionof{figure}{Confidence interval results for $m_2$, $N=4$}
   \label{fig:ci_vc_p3}
\end{minipage}
\vspace{-0.7cm}
\end{figure}

\noindent We discuss here some hurdles we encountered with the application of \fact{}. 

\paragraph{Scalability.} Technically, even for a DTMC transition that has zero probability mass due to zero observations in the data, for the purpose of confidence analysis, we should add a parameter to the model representing the probability of the transition (and specify the number of observations as zero). However, we found this approach scales poorly. 
For instance, for model $m_1$ and \emph{Property 2} (which is the easiest to analyze), we could not obtain results beyond $N=3$ with \fact{} and we could not obtain any results for the other properties or for $m_2$. 

In contrast, if we remove the transitions with zero observations, the model has fewer parameters and the analysis scales better. For instance, on a Windows OS machine with 11th Gen Intel(R) Core(TM) i7 processor and 64GB RAM,  \fact{} finishes in 5.783s (model without zero-observations transitions) vs. 140.676s (model with zero-observations transitions).   

Figure~\ref{fig:CI_0obs} shows the results obtained for $m_1$ with and without modeling transitions with zero observations. 
As the difference between computed intervals is very small,  we have decided to report in this paper on the analysis for models with those transitions removed (for which we could obtain results for $N=4$).
Note that whether these transitions are included or not has no effect on the results of the \prism{} analysis. Also note that one could attempt to  prove the absence of such transitions (using DNN-specific analysis) obtaining as a result an abstraction that is more amenable to verification.

Scalability issues in \fact{} arise due to the complexity of the computational problems solved by \fact{} in each of its two stages. In the first stage, \fact{} uses {\em parametric model checking} to generate an algebraic expression (for the probability of the property $\phi$ under analysis) which is a multivariate rational function that depends on the unknown transition probabilities. This stage can lead to scalability issues due to state explosion during parametric model checking. In the second stage, referred to {\em confidence interval inference}, \fact{} uses the observations to derive confidence intervals for each unknown transition probability appearing in the algebraic expression and then uses them to derive the required confidence interval for $\phi$; the analysis is based on the concept of simultaneous confidence intervals for a multinomial distribution and requires solving optimization problems. The second stage can give rise to scalability issues due the complexity of the optimization problems to be solved. Newer work~\cite{Fang22CGA} promises to scale much better, however we have found it not mature enough to be applicable to our models. We are working on improving the tool for future use.

%Scalability issues in \fact{} are due to the state explosion during parametric model checking and the complexity of the optimization problems to be solved during the confidence interval inference stage. Newer work~\cite{Fang22CGA} promises to scale much better, however we have found it not mature enough to be applicable to our models. We are working on improving the tool for future use.

\paragraph{Numerical Issues.} We conducted confidence analysis for \emph{Property 3} on model $m_2$, for $N=4$, without the zero-observations transitions. Figure~\ref{fig:ci_vc_p3} shows the calculated confidence intervals for the probability of satisfying the property at different confidence levels. One would expect, (i) the intervals to get larger for higher confidence levels; (ii) the intervals at higher confidence levels to include the ones at lower levels. However, the probabilities for \emph{Property 3} are very small (on the order of $1e$-6). We suspect that the Gurobi Optimizer~\cite{gurobi} (which FACT uses internally at the confidence interval inference stage) is unable to handle the problem of finding the minima and maxima corresponding to the confidence interval bounds for such small probabilities, and as a consequence, we see confidence intervals that do not grow monotonically in size with the confidence levels.
%This leads to numerical precision issues in the \fact{} calculations, particularly at the confidence interval inference stage where \fact{} solves optimization problems by internally invoking the Gurobi optimizer~\cite{gurobi}.
Additionally, the confidence intervals being calculated are meant to be conservative. As such, having non-monotonically growing intervals with increasing confidence levels is not necessarily wrong. It may only mean that the analysis engine could not do any better, and the intervals at lower confidence levels are over-conservative.

\end{document}